\documentclass[referee,sn-basic]{sn-jnl}

\usepackage{graphicx}%
\usepackage{subcaption}
\usepackage{caption}
\usepackage{multirow}%
\usepackage{amsmath,amssymb,amsfonts}%
\usepackage{amsthm}%
\usepackage{mathrsfs}%
\usepackage[title]{appendix}%
\usepackage{xcolor}%
\usepackage{textcomp}%
\usepackage{manyfoot}%
\usepackage{booktabs}%
\usepackage{algorithm}%
\usepackage{algorithmicx}%
\usepackage{algpseudocode}%
\usepackage{listings}% 

\theoremstyle{thmstyleone}%
\theoremstyle{thmstyletwo}%

\theoremstyle{thmstylethree}%
\newtheorem{definition}{Definition}%

\begin{document}

\title[Article Title]{Communication-Aware Asynchronous Distributed Trajectory Optimization for UAV Swarm}

\author[1]{\fnm{Yue} \sur{Yu}}

\author[2]{\fnm{Xiaobo} \sur{Zheng}}

\author*[1]{\fnm{Shaoming} \sur{He}}\email{shaoming.he@bit.edu.cn}

\affil[1]{\orgdiv{School of Aerospace Engineering}, \orgname{Beijing Institute of Technology}, \orgaddress{\street{5 South Zhongguancun Street}, \city{Beijing}, \postcode{100081}, \country{China}}}

\affil[2]{\orgname{Beijing Institute of Aerospace Systems Engineering}, \orgaddress{\street{Street}, \city{Beijing}, \postcode{100076}, \country{China}}}

\abstract{Distributed optimization offers a promising paradigm for trajectory planning in Unmanned Aerial Vehicle (UAV) swarms, yet its deployment in communication-constrained environments remains challenging due to unreliable links and limited data exchange. This paper addresses this issue via a two-tier architecture explicitly designed for operation under communication constraints. We develop a Communication-Aware Asynchronous Distributed Trajectory Optimization (CA-ADTO) framework that integrates Parameterized Differential Dynamic Programming (PDDP) for local trajectory optimization of individual UAVs with an asynchronous Alternating Direction Method of Multipliers (async-ADMM) for swarm-level coordination. The proposed architecture enables fully distributed optimization while substantially reducing communication overhead, making it suitable for real-world scenarios in which reliable connectivity cannot be guaranteed. The method is particularly effective in handling nonlinear dynamics and spatio-temporal coupling under communication constraints.}


\maketitle

\section{Introduction}\label{sec1}

UAV swarms have emerged as transformative systems for complex missions including wildfire surveillance \citep{bib1}, intelligence surveillance and reconnaissance \citep{bib2}, situational awareness \citep{bib3}, and cooperative interception \citep{bib4}. In these applications, trajectory optimization is the cornerstone for ensuring both mission success and operational safety \citep{sezer2022optimized,qian2020improved,sanchez2020trajectory}. Over the past decade, trajectory optimization techniques have evolved from sophisticated single-agent formulations to distributed multi-agent frameworks, driven by the increasing scale and complexity of swarm-based missions \citep{saravanos2023distributed}.

For individual UAV trajectory optimization, a variety of numerical methods have demonstrated strong performance. Pseudospectral methods achieve high-accuracy solutions by discretizing continuous-time problems \citep{bib5}, while sequential quadratic programming (SQP) \citep{bib6} and sequential convex programming (SCP) \citep{bib7} provide flexible tools for handling nonlinear dynamics and constraints. Among these approaches, differential dynamic programming (DDP) has gained particular prominence due to its rapid convergence and solid theoretical foundation. By leveraging second-order Taylor expansions and Bellman’s optimality principle, DDP effectively mitigates the curse of dimensionality \citep{bib8,bib9,bib10,bib11}.

Extending trajectory optimization from single- to multi-agent systems introduces fundamental architectural challenges. Centralized optimization offers conceptually straightforward solutions but suffers from single points of failure and limited scalability \citep{bib12}. Distributed architectures, in contrast, enhance robustness and scalability, and have given rise to three main classes of methods: distributed gradient descent \citep{bib13}, distributed SCP \citep{bib14,bib15}, and the alternating direction method of multipliers (ADMM) \citep{bib16,bib17}. Among these, ADMM has proven particularly attractive because of its decomposability and convergence guarantees \citep{bib18,bib19}. By decomposing large-scale problems into smaller, parallelizable subproblems, ADMM is especially well suited to multi-agent systems \cite{bib16}. Building on this, the distributed parameterized differential dynamic programming (D-PDDP) algorithm combines the decomposition capability of ADMM with the optimization power of DDP, resulting in a fully decentralized framework capable of handling complex spatio-temporal constraints while maintaining collision avoidance and temporal consensus \citep{bib20}.

However, most existing ADMM-based methods, including D-PDDP, are developed under idealized communication assumptions, such as synchronous updates over lossless, delay-free links. These assumptions are rarely satisfied in real-world UAV swarm operations \citep{bib20}. A notable research gap therefore lies in adapting ADMM-based distributed optimization to severely communication-constrained environments, where packet loss, delays, intermittent connectivity, and limited bandwidth are prevalent \citep{bib21,bib22}. In large-scale swarms, these issues are further exacerbated by dynamic network topologies and contention for scarce communication resources among many agents. Under such conditions, enforcing synchronous updates can create performance bottlenecks that severely degrade coordination quality and limit system scalability \citep{bib23}. Asynchronous ADMM (async-ADMM) has been introduced to alleviate the straggler problem and improve robustness under communication delays by allowing agents to update their variables without waiting for all peers, thereby enhancing system responsiveness \citep{bib24}.

Motivated by these challenges and the lack of communication-aware distributed optimization frameworks for UAV swarms, this paper proposes a distributed spatio-temporal trajectory optimization method based on Parameterized DDP (PDDP) and async-ADMM. The algorithm adopts a two-level architecture: at the lower level, the PDDP layer solves local trajectory optimization problems for each UAV subject to nonlinear dynamics and control limits; at the upper level, the async-ADMM layer enforces local path constraints and achieves spatio-temporal coordination among all agents, with particular emphasis on operation under constrained communication resources. The main contribution of this work is a distributed trajectory optimization framework that explicitly incorporates asynchronous communication in a swarm of agents operating under limited communication capacity. A series of numerical simulations with flight-time optimization and time coordination constraints is carried out to evaluate the effectiveness of the proposed method. To the best of our knowledge, no comparable communication-aware asynchronous distributed trajectory optimization framework for UAV swarms has been reported in the literature. The results show that the proposed algorithm can handle various mission scenarios—such as free terminal time, simultaneous arrival, and specified time intervals—and can be applied to large-scale UAV swarms.

The remainder of this paper is organized as follows. Section \ref{sec2} introduces the general UAV swarm trajectory optimization problem and reformulates it for distributed computation. Section \ref{sec3} details the proposed CA-ADTO algorithm. Section \ref{sec4} presents numerical simulations that evaluate the algorithm’s performance, and Section \ref{sec5} concludes the paper.

\section{Backgrounds and Preliminaries}\label{sec2}

This section introduces the general formulation of the UAV Swarm Trajectory Optimization (USTO) problem and reformulates it into a form amenable to distributed optimization.

\subsection{UAV Swarm Trajectory Optimization Problem}\label{sec2:sub1}

Consider a swarm of $M$ agents, indexed by the set $\mathcal{M}=\{1,\ldots,M\}$. Before presenting the formal optimization problem, we define the neighborhood relationships among agents following \cite{bib9}.

\begin{definition}[Neighbor set]
For each agent $i\in\mathcal{M}$, its neighbor set $\mathcal{N}_i\subseteq\mathcal{M}$ contains all agents $j$ that are within the effective communication range of $i$, including $i$ itself.
\end{definition}
\begin{definition}[Deemed Neighbor Set]
The deemed neighbor set $\mathcal{P}_i = \{ j \in \mathcal{M} \mid i \in \mathcal{N}_j \}$ consists of all agents $j$ that regard agent $i$ as their neighbor.
\end{definition}

Note that neighborhood relations need not be symmetric; that is, $j \in \mathcal{N}_i$ does not necessarily imply $i \in \mathcal{N}_j$. Communication among UAVs is assumed to be local under the condition that each agent $i \in \mathcal{M}$ can exchange information only with agents $j \in \mathcal{N}_i \cap \mathcal{P}_i$ that mutually regard one another as neighbors. In a discrete-time setting, the nonlinear dynamics of each agent $i$ are described by
\begin{equation}\boldsymbol{x}_{i,k+1}=\boldsymbol{f}_i(\boldsymbol{x}_{i,k},\boldsymbol{u}_{i,k})\label{eq1}\end{equation}
where $\boldsymbol{x}_{i,k}\in\mathbb{R}^{p_i}$ and $\boldsymbol{u}_{i,k}\in\mathbb{R}^{q_i}$ respectively represent the state and control input vectors of the $i$-th agent at time step $k$. The subscript $k=\{0,1,...,N\}$ represent the time instant in the discrete framework, and the value of a variable at time instant $t_{k}$ is indicated by the variable with subscript $k$. $\boldsymbol{f}_{i}$ represents the system's discrete, nonlinear dynamic properties. 

The collective objective of the multi-agent system is to minimize a global cost function, defined as the sum of individual local cost functions 
\begin{equation}J=\sum_{i=1}^MJ_i(\boldsymbol{X}_i,\boldsymbol{U}_i)\label{eq2}
\end{equation}
in which $\boldsymbol{X}_{i}$ and $\boldsymbol{U}_{i}$ are the aggregated state and control vectors, respectively, for agent $i$ across all time instants. Each agent's local objective function $J_i(\boldsymbol{X}_i,\boldsymbol{U}_i)$ comprises a running cost and a terminal cost, formalized as 
\begin{equation}J_i(\boldsymbol{X}_i,\boldsymbol{U}_i)=\sum_{k=0}^{N-1}l_i(\boldsymbol{x}_{i,k},\boldsymbol{u}_{i,k})+\phi_i(\boldsymbol{x}_{i,N},t_{i,N})\label{eq3}\end{equation}
in which the scalar function $l_i(\boldsymbol{x}_{i,k},\boldsymbol{u}_{i,k})$ represents the running cost at each time step, while $\phi_i(\boldsymbol{x}_{i,N},t_{i,N})$ represents the cost evaluated at the terminal state and time. Owing to physical actuation limits, the control input for each agent should satisfy a set of constraints, 
\begin{equation}\boldsymbol{b}_{i,k}(\boldsymbol{u}_{i,k})\leq0\label{eq4}\end{equation}
where $\boldsymbol{b}_{i,k}(\boldsymbol{u}_{i,k})$ represents the constraints on the control effort of agent $i$ at time instant $k$. 

In addition to constraints on control effort, each agent should also adhere to nonlinear path constraints such as obstacle avoidance
\begin{equation}\boldsymbol{a}_{i,k}(\boldsymbol{x}_{i,k})\leq0\label{eq5}
\end{equation}
where $\boldsymbol{a}_{i,k}(\boldsymbol{x}_{i,k})$ is a nonlinear function for describing path constraints. 

In spatial-temporal optimization problems, the terminal time for each agent is not fixed a priori but is treated as a free optimization variable. However, this variable should lie within the feasible range attainable by the agent's dynamics. This terminal time constraint is generically formulated as 
\begin{equation}c_{i,N}(t_{i,N})\leq0\label{eq6}\end{equation}
where $c_{i,N}(t_{i,N})$ is a general nonlinear function that bounds the terminal time. 

To capture inter-agent coupling, we introduce spatial and temporal constraints between each agent $i$ and its neighbors $j \in \mathcal{N}_i \setminus \{i\}$. The inter-agent state constraint is given by
\begin{equation}\boldsymbol{d}_{ij,k}(\boldsymbol{x}_{i,k},\boldsymbol{x}_{j,k})\leq0\label{eq7}\end{equation}
and the inter-agent terminal time constraint is
\begin{equation}\boldsymbol{e}_{ij,N}(t_{i,N},t_{j,N})\leq0\label{eq8}\end{equation}
where $\boldsymbol{d}_{ij,k}(\boldsymbol{x}_{i,k},\boldsymbol{x}_{j,k})$ typically enforces safety or connectivity conditions (e.g., collision avoidance or maintaining communication range), and $\boldsymbol{e}_{ij,N}(t_{i,N},t_{j,N})$ coordinates the mission completion times (e.g., simultaneous arrival or a prescribed sequence).

Practical swarm operations must also cope with unreliable communication and limited data exchange. To model link reliability, we introduce a Bernoulli random variable $L_{ij,k}$ representing the success ($1$) or failure ($0$) of a transmission from agent $i$ to agent $j$ at time step $k$ as
\begin{equation}P(L_{ij,k}=1)=p_{ij,k}\label{eq9}
\end{equation}
where $p_{ij,k} \in [0,1]$ is the communication success probability, encapsulating effects such as high traffic load and platform limitations.

Summarizing the above components, the $USTO_1$ addressed in this paper can be formulated as

$USTO_1$: Find
\begin{equation}\begin{aligned}
\{\boldsymbol{U}_i^*,t_{i,N}^*\} & =\min\sum_{i=1}^MJ_i\left(\boldsymbol{X}_i,\boldsymbol{U}_i\right) \\
s.t.\quad & \boldsymbol{x}_{i,k+1}=\boldsymbol{f}_i(\boldsymbol{x}_{i,k},\boldsymbol{u}_{i,k}) \\
 & \boldsymbol{a}_{i,k}(\boldsymbol{x}_{i,k})\leq0 \\
 & \boldsymbol{b}_{i,k}(\boldsymbol{u}_{i,k})\leq0 \\
 & c_{i,N}(t_{i,N})\leq0 \\
 & \boldsymbol{d}_{ij,k}(\boldsymbol{x}_{i,k},\boldsymbol{x}_{j,k})\leq0, \quad j\neq i \\
 & \boldsymbol{e}_{ij,N}(t_{i,N},t_{j,N})\leq0, \quad j\neq i \\
 & P(L_{ij,k}=1)=p_{ij,k}, \quad p_{ij,k}\in[0,1]
\end{aligned}\label{eq10}\end{equation}
for all $i \in \mathcal{M}$ and $k = 0,1,\dots,N-1$. The nonlinearities in both dynamics and constraints typically render closed-form solutions intractable; thus, $USTO_1$ must be solved via numerical optimization methods.

\subsection{Problem Reformulation}\label{sec2:sub2}
To solve the general multi-agent problem \eqref{eq10} using distributed optimization \citep{bib4}, we first reformulate it into a structure compatible with the ADMM framework. To this end, we introduce `safe' copy variables $\tilde{\boldsymbol{x}}_{i,k}\in\mathbb{R}^{p_i}$, $\tilde{\boldsymbol{u}}_{i,k}\in\mathbb{R}^{q_i}$, $\tilde{t}_{i,N}\in\mathbb{R}$ of $\boldsymbol{x}_{i,k}$, $\boldsymbol{u}_{i,k}$ and $t_{i,N}$ for each agent $i$, which are constrained to be consistent with the primal variables generated by the PDDP layer
\begin{equation}\tilde{\boldsymbol{x}}_{i,k}=\boldsymbol{x}_{i,k},\tilde{\boldsymbol{u}}_{i,k}=\boldsymbol{u}_{i,k},\tilde{t}_{i,N}=t_{i,N}\label{eq11}\end{equation}
and are simultaneously required to satisfy the agent's local constraints, i.e., $\boldsymbol{a}_{i,k}(\tilde{\boldsymbol{x}}_{i,k})\leq0$, $\boldsymbol{b}_{i,k}(\tilde{\boldsymbol{u}}_{i,k})\leq0$ and $c_{i,N}(\tilde{t}_{i,N})\leq0$. 

The inter-couplings from constraints (\ref{eq7}) and (\ref{eq8}) prevent a decentralized solution to Problem (\ref{eq10}). To resolve this, the proposed architecture requires each agent to keep a `safe' copy of its own and its neighbors' variables to achieve consensus. Specifically, from the perspective of agent $i$, the state and terminal time of a neighbor $j$ are represented by the variables $\widetilde{\boldsymbol{x}}_{j,k}^i\in\mathbb{R}^{p_i}$, $\tilde{t}_{j,N}^i\in\mathbb{R},j\in\mathcal{N}_i,i\in\mathcal{M}$, where $\tilde{\boldsymbol{x}}_{j,k}^i$ is a local `safe' copy and $\tilde{t}_{j,N}^i$ is a proper final time of agent $j$ from agent $i$’s perspective. Consequently, we define the augmented state and time vectors for agent $i$ by stacking the corresponding variables from all its neighbors 
\begin{equation}\tilde{\boldsymbol{x}}_{i,k}^a=\left[\left\{\tilde{\boldsymbol{x}}_{j,k}^i\right\}_{j\in\mathcal{N}_i}\right]\in\mathbb{R}^{\widetilde{p}_i},\widetilde{p}_i=\sum_{j\in\mathcal{N}_i}p_j\label{eq12}\end{equation}

\begin{equation}\tilde{t}_{i,N}^a=\left[\left\{\tilde{t}_{j,N}^i\right\}_{j\in\mathcal{N}_i}\right]\in\mathbb{R}^{|\mathcal{N}_i|}\label{eq13}\end{equation}

To decentralize the problem, constraints (\ref{eq7}) and (\ref{eq8}) should be formulated from the local perspective of each agent $i$ using the introduced copy variables. Consequently, the inter-agent state constraint is reformulated as $\boldsymbol{d}_{ij,k}(\tilde{\boldsymbol{x}}_{i,k},\tilde{\boldsymbol{x}}_{j,k}^i)\leq0$ and the inter-agent time constraint as $\boldsymbol{e}_{ij,N}(\tilde{t}_{i,N},\tilde{t}_{j,N}^i)\leq0$. These can also be expressed in a compact form with 
\begin{equation}\boldsymbol{d}_{i,k}^a(\widetilde{\boldsymbol{x}}_{i,k}^a)\leq0, \boldsymbol{e}_{i,N}^a(\widetilde{t}_{i,N}^a)\leq0\label{eq14}\end{equation}
in which $\boldsymbol{d}_{i,k}^a(\widetilde{\boldsymbol{x}}_{i,k}^a)=\left[\left\{\boldsymbol{d}_{ij,k}(\widetilde{\boldsymbol{x}}_{i,k},\widetilde{\boldsymbol{x}}_{j,k}^i)\right\}_{j\in\mathcal{N}_i\setminus\{i\}}\right]$, $\boldsymbol{e}_{i,N}^a(\tilde{t}_{i,N}^a)=\left[\left\{\boldsymbol{e}_{ij,N}(\tilde{t}_{i,N},\tilde{t}_{j,N}^i)\right\}_{j\in\mathcal{N}_i\setminus\{i\}}\right]$.

Furthermore, to reconcile the discrepancies that arise when multiple neighbors of agent $i$ maintain differing copies of its variables, a consensus should be enforced. This is achieved by introducing global consensus variables $\boldsymbol{z}_{i,k}\in\mathbb{R}^p$ and $\boldsymbol{s}_{iN}\in\mathbb{R}$ for the state and terminal time, respectively, where $i\in\mathcal{M}$. The consensus constraints 
\begin{equation}\tilde{\boldsymbol{x}}_{j,k}^i=\boldsymbol{z}_{j,k},\tilde{t}_{j,N}^i=s_{j,N},j\in\mathcal{N}_i\label{eq15}\end{equation}
require all local perspectives to align with these global variables. A compact form of (\ref{eq15}), considering all neighbors of agent $i$ is given by 
\begin{equation}\tilde{\boldsymbol{x}}_{i,k}^a=\boldsymbol{z}_{i,k}^a,\tilde{t}_{i,N}^a=s_{i,N}^a\label{eq16}\end{equation}
where $\boldsymbol{z}_{i,k}^\mathrm{a}=\left[\left\{\boldsymbol{z}_{j,k}\right\}_{j\in\mathcal{N}_i}\right]\in\mathbb{R}^{\widetilde{p}_i}$, $s_{i,N}^\mathrm{a}=\left[\left\{s_{j,N}\right\}_{j\in\mathcal{N}_i}\right]\mathbb{R}^\mathcal{M}$. 

Equations (\ref{eq11})–(\ref{eq16}) can be solved using the standard ADMM. However, due to unreliable communication links during swarm trajectory optimization, as modeled in Eq. (\ref{eq9}), UAVs within the swarm cannot perform real-time synchronous information updates, necessitating an asynchronous distributed computation framework. To address this, we introduce a dual-role architecture composed of \textit{master} and \textit{worker}. In this structure, the master acts as a virtual node responsible for updating the global consensus variables, while each UAV in the swarm functions as a worker. Each worker minimizes its local objective $\boldsymbol{f}_i$ (in parallel) based on its data subset; and sends the updated local primal variables and local dual variables to the master. The master, in turn, updates global variables by driving the local variables into consensus, and then distributes the updated value back to the workers, and the process re-iterates. Due to unreliable communication links in inter-UAV communications, the transmission of local variables from individual workers to the master node is subject to corresponding inaccuracies. We define a \textit{successful communication} as the event where a worker correctly transmits its local variables to the master. The probability of successful communication is given by:
\begin{equation}P(L_{w-m,k}=1)=p_{con}\label{eq17}
\end{equation}
where the probability of successful communication between worker and master is reformulated as $L_{w-m,k}$ ,and $p_{con}\in[0,1]$ is the swarm’s \textit{communication connection probability}, which encapsulates the effects of high traffic load and platform limitations.

Therefore, the original $USTO_1$ problem can be reformulated into the following equivalent optimization problem. 

$USTO_2$: Find
\begin{equation}\begin{gathered}
\{\boldsymbol{U}_i^*,t_{i,N}^*\}=\min\sum_{i=1}^MJ_i\left(\boldsymbol{X}_i,\boldsymbol{U}_i\right) \\
s.t.\quad\boldsymbol{x}_{i,k+1}=\boldsymbol{f}_i(\boldsymbol{x}_{i,k},\boldsymbol{u}_{i,k}) \\
\boldsymbol{a}_{i,k}(\widetilde{\boldsymbol{x}}_{i,k})\leq0 \\
\boldsymbol{b}_{i,k}(\widetilde{\boldsymbol{u}}_{i,k})\leq0 \\
c_{i,N}(\tilde{t}_{i,N})\leq0 \\
\boldsymbol{d}_{i,k}^a(\widetilde{\boldsymbol{x}}_{i,k}^a)\leq0 \\
\boldsymbol{e}_{i,N}^a(\tilde{\boldsymbol{t}}_{i,N}^a)=0 \\
P(L_{w-m,k}=1)=p_{con}, \quad p_{con}\in[0,1] \\
\boldsymbol{x}_{i,k}=\widetilde{\boldsymbol{x}}_{i,k},\boldsymbol{u}_{i,k}=\widetilde{\boldsymbol{u}}_{i,k},t_{i,N}=\widetilde{t}_{i,N},\widetilde{\boldsymbol{x}}_{i,k}^{a}=\boldsymbol{z}_{i,k}^{a},\widetilde{t}_{i,N}^{a}=s_{i,N}^{a}
\end{gathered}\label{eq18}\end{equation}
for all $i \in \mathcal{M}$ and $k = 0,1,\dots,N-1$. This formulation explicitly separates local dynamics and constraints from global consensus, paving the way for a communication-aware asynchronous ADMM solution.

\subsection{Schematic diagram of a two-tiered architecture for CA-ADTO}\label{sec2:sub3}

\begin{figure}[h]
\centering
\includegraphics[width=0.7\textwidth]{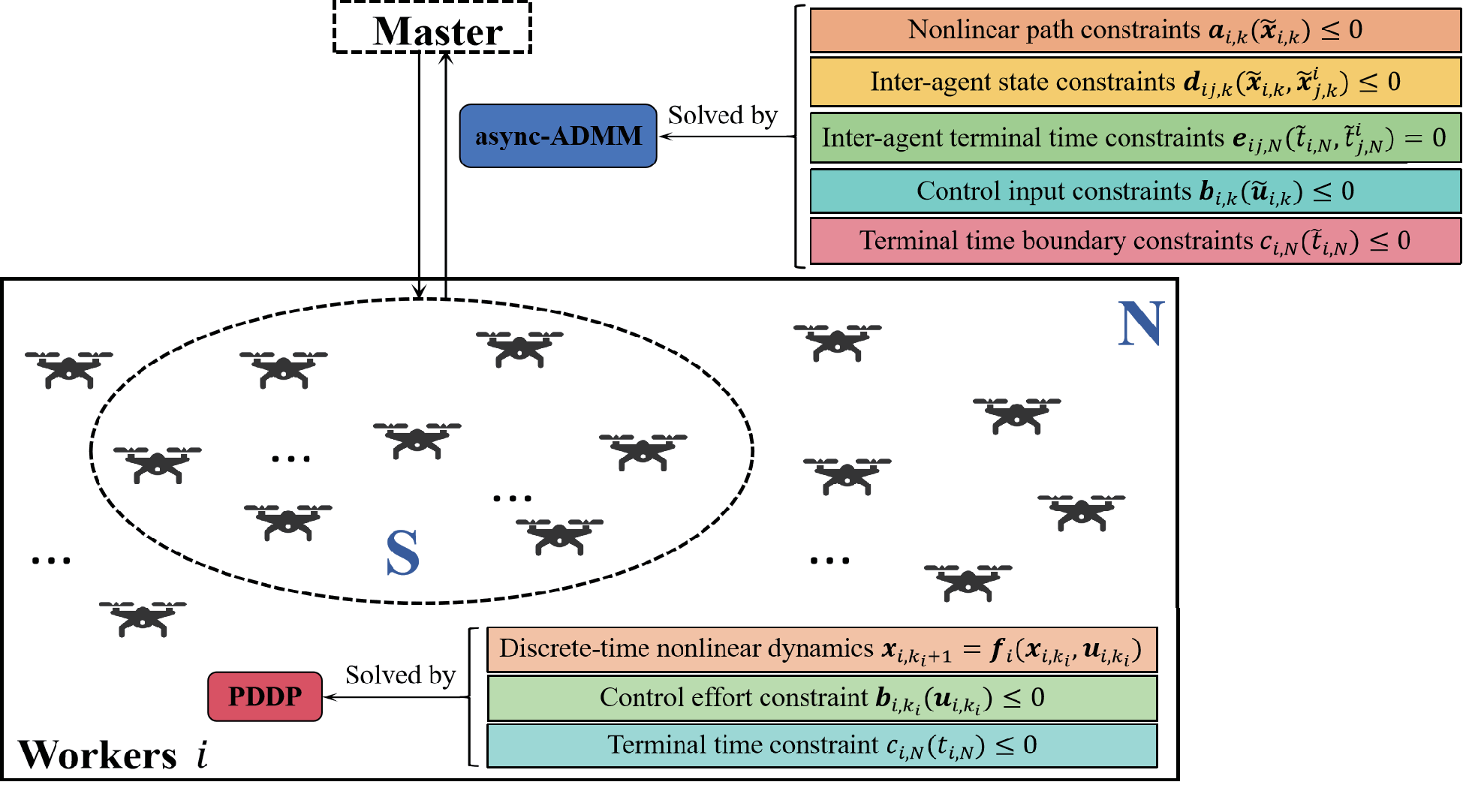}
\caption{Schematic diagram of a two-tiered architecture for CA-ADTO.}\label{fig1}
\end{figure}

The algorithm proposed in this paper to solve $USTO_2$ has a two-tiered architecture. In this structure, the async-ADMM layer manages inter-agent spatiotemporal constraints to achieve swarm consensus, while the PDDP layer handles the free terminal time trajectory optimization for master and workers, accounting for their local dynamics and control constraints. The cooperative framework, illustrated in Figure \ref{fig1}, enables efficient distributed planning: PDDP acts as the local trajectory optimizer for each UAV, while async-ADMM coordinates these local solutions to satisfy global collaborative constraints.

There are two essential ingredients for async-ADMM. (i) Instead of requiring full synchronization on all the workers in each ADMM iteration, a partial synchronization is only needed. (ii) While updates from the faster workers will be incorporated more often by the master, we require that updates from the slow workers cannot be older than a certain maximum delay. As for the async-ADMM, the master is responsible for updating the consensus variables, while each worker $i$ is responsible for updating the local primal variables and local dual variables. As the proposed algorithm is fully asynchronous, the master keeps a clock $k$, which starts from zero and is incremented by 1 after each global variable update. Similarly, every worker also has its own clock $k_i$, which starts from zero and is incremented by 1 after each dual variable update. All the clocks $k$ and $\{k_i\}_{i=1}^N$ are run independently.

The master waits for the workers’local variables update before it can update global variables. Recall that for the synchronous ADMM, this can proceed only after the $\{\boldsymbol{x}_i\}$ updates from all $N$ workers have finished. In distributed systems, this mechanism is called a barrier, and is the simplest synchronization primitive. However, it suffers from the straggler problem and allows the system to move forward only at the pace of the slowest worker. To alleviate this problem, we relax it to a \textit{partial barrier}. Specifically, the master only needs to wait for a minimum of $S$ updates, where $S(\geq1)$ can be much smaller than $N$. The synchronous ADMM can be regarded as using the extreme setting of $S=N$. Moreover, recall that some workers are using out-of-dateversions of global variables. Consequently, their local variable updates are also out-of-date, an issue that the master has to cope with.

Besides, the master needs to wait for another precondition to be satisfied before it can proceed. Note that if we only rely on a partial barrier with small $S$, updates from the slow workers will be incorporated into global variables much less often than those from the faster workers. To ensure sufficient freshness of all the updates, we enforce a \textit{bounded delay} condition. Specifically, update from every worker has to be serviced by the master at least once every $\tau$ iterations, where $\tau\geq1$ is a user-defined parameter.

When both the partial barrier and bounded delay conditions are met, the master can proceed with the global variables update. Note that the diagram is a schematic representation illustrating quantitative relationships among UAVs, not their physical positions. Specifically, the communication links between workers and the master represent logical connectivity, which is independent of their spatial proximity.

\section{Communication-Aware Asynchronous Distributed Trajectory Optimization}\label{sec3}

This section provides a comprehensive explanation of CA-ADTO algorithm. Our algorithm is derived based on the async-ADMM framework. Within the context of free terminal time spatial-temporal trajectory optimization, the terminal time $t_{i,N}$ of each agent is selected as the unknown parameter $\boldsymbol{\theta}$ in the PDDP formulation. Consequently, the symbol $t_{i,N}$ serves a dual purpose in the subsequent derivation: it represents the core variable in the spatial-temporal optimization and, simultaneously, denotes the key unknown parameter being optimized by the PDDP algorithm in this section.

To Problem (\ref{eq18}) using async-ADMM, we first recast it into the standard ADMM form. This is achieved by incorporating the constraints into the objective function using corresponding indicator functions  
\begin{equation}\begin{aligned}
&\min\sum_{i=1}^M\sum_{k=0}^NJ_i\left(\boldsymbol{x}_{i,k},\boldsymbol{u}_{i,k}\right)+I_{\boldsymbol{f}_i}(\boldsymbol{x}_{i,k},\boldsymbol{u}_{i,k})+I_{\boldsymbol{b}_{i,k}}(\widetilde{\boldsymbol{u}}_{i,k})+I_{C_i}(\widetilde{t}_{i,N})
+I_{\boldsymbol{a}_{i,k}}(\widetilde{\boldsymbol{x}}_{i,k},\widetilde{\boldsymbol{u}}_{i,k})+I_{\boldsymbol{d}_{i,k}^\alpha}(\widetilde{\boldsymbol{x}}_{i,k}^\alpha)+I_{\boldsymbol{e}_{i,N}^\alpha}(\widetilde{\boldsymbol{t}}_{i,N}^\alpha)
\\
&s.t.\quad\boldsymbol{u}_{i,k}=\tilde{\boldsymbol{u}}_{i,k},\boldsymbol{x}_{i,k}=\tilde{\boldsymbol{x}}_{i,k},t_{i,N}=\tilde{t}_{i,N},\tilde{\boldsymbol{x}}_{i,k}^a=\boldsymbol{z}_{i,k}^a,\tilde{\boldsymbol{t}}_{i,N}^a=\boldsymbol{s}_{i,N}^a
\label{eq19}\end{aligned}\end{equation}
where $J_i(\boldsymbol{x}_{i,k},\boldsymbol{u}_{i,k})=l_i(\boldsymbol{x}_{i,k},\boldsymbol{u}_{i,k})$, if $k<N$ and $J_i(\boldsymbol{x}_{i,k},\boldsymbol{u}_{i,k})=\phi_i(\boldsymbol{x}_{i,N},t_{i,N})$ if $k=N$. The indicator function $I_C(\boldsymbol{x})$ is defined to denote whether the vector $\boldsymbol{x}$ is on the set $C$. If the vector $\boldsymbol{x}$ is on the set $C$, then $I_C(\boldsymbol{x})=0$; conversely $\boldsymbol{x}$ is not on $C$, then $I_C(x)=+\infty$. 

Following the standard ADMM derivation, we now construct the augmented Lagrangian for Problem (\ref{eq19}) 
\begin{equation}\begin{aligned}
  \mathscr{L}  =&\sum_{i=1}^M\sum_{k=0}^NJ_i\left(\boldsymbol{x}_{i,k},\boldsymbol{u}_{i,k}\right)+I_{\boldsymbol{f}_i}(\boldsymbol{x}_{i,k},\boldsymbol{u}_{i,k})+I_{\boldsymbol{b}_{i,k}}(\widetilde{\boldsymbol{u}}_{i,k})+I_{\boldsymbol{c}_i}(\widetilde{t}_{i,N}) \\
 &  +I_{\boldsymbol{a}_{i,k}}(\widetilde{\boldsymbol{x}}_{i,k})+I_{\boldsymbol{d}_{i,k}^{a}}(\widetilde{\boldsymbol{x}}_{i,k}^{a})+I_{\boldsymbol{e}_{i,k}^{a}}(\widetilde{\boldsymbol{t}}_{i,N}^{a}) \\
 &  +\boldsymbol{\zeta}_{i,k}^T(\boldsymbol{u}_{i,k}-\widetilde{\boldsymbol{u}}_{i,k})+\boldsymbol{\lambda}_{i,k}^T(\boldsymbol{x}_{i,k}-\widetilde{\boldsymbol{x}}_{i,k})+\nu_{i,N}^T(t_{i,N}-\widetilde{t}_{i,N}) \\
 &  +\boldsymbol{y}_{i,k}^T(\widetilde{\boldsymbol{x}}_{i,k}^a-\boldsymbol{z}_{i,k}^a)+\boldsymbol{\eta}_{i,N}^T(\widetilde{t}_{i,N}^a-s_{i,N}^a) \\
 &  +\frac{\tau}{2}\|\boldsymbol{u}_{i,k}-\widetilde{\boldsymbol{u}}_{i,k}\|_2^2+\frac{\rho}{2}\|\boldsymbol{x}_{i,k}-\widetilde{\boldsymbol{x}}_{i,k}\|_2^2+\frac{\sigma}{2}\|t_{i,N}-\widetilde{t}_{i,N}\|_2^2 \\
 &  +\frac{\mu}{2}\|\tilde{\boldsymbol{x}}_{i,k}^a-\boldsymbol{z}_{i,k}^a\|_2^2+\frac{\gamma}{2}\|\tilde{t}_{i,k}^a-s_{i,N}^a\|_2^2
\label{eq20}\end{aligned}\end{equation}
where the dual variables of constraints are represented by $\boldsymbol{\zeta}_{i,k}$, $\boldsymbol{\lambda}_{i,k}$, $\nu_{i,N}$, $\boldsymbol{y}_{i,k}$, $\boldsymbol{\eta}_{i,N}$, and the penalty parameters of constraints are denoted using $\tau,\rho,\sigma,\mu,\gamma>0$. As for the async-ADMM, the master keeps a clock $k$, every worker also has its own clock $k_i$. Let  $\boldsymbol{x}_{i,k_i}$, $\boldsymbol{\zeta}_{i,k}$, $\boldsymbol{\lambda}_{i,k}$, $\boldsymbol{y}_{i,k}$ be the values of $\boldsymbol{x}_i$ and $\boldsymbol{\zeta}_i$, $\boldsymbol{\lambda}_i$, $\boldsymbol{y}_i$ when worker $i$’s clock is at $k_i$; and $\boldsymbol{z}_k$ be the value of $\boldsymbol{z}$ when the master’s clock is at $k$. The CA-ADTO is divided into four optimization steps, which are derived as follows.

\subsection{Optimization Steps 1: Single-Agent PDDP Update}\label{sec3:sub1}

We first consider a particular worker $i$ (at time $k_i$). Using the most recent $\boldsymbol{z}$, $s$ value received by $i$ from the master.

In this update step, the augmented Lagrangian is minimized with respect to the local variables $\boldsymbol{x}_{i,k_i}$, $\boldsymbol{u_{i,k_i}}$, $t_{i,N}$ using the PDDP algorithm, which serves as a single-agent spatial-temporal trajectory optimizer, i.e., 
\begin{equation}\left\{\boldsymbol{x}_{i,k_i},\boldsymbol{u}_{i,k_i},t_{i,N}\right\}^{n+1}=\mathrm{argmin}\mathscr{L}\left(\boldsymbol{x}_{i,k_i},\boldsymbol{u}_{i,k_i},t_{i,N},\widetilde{\boldsymbol{x}}_{i,k_i}^{a,n},\widetilde{\boldsymbol{u}}_{i,k_i}^n,\widetilde{\boldsymbol{t}}_{i,N}^{a,n},\boldsymbol{z}_{i,k_i}^{a,n},s_{i,N}^{a,n},\boldsymbol{\zeta}_{i,k_i}^n,\boldsymbol{\lambda}_{i,k_i}^n,\nu_{i,N}^n,\boldsymbol{y}_{i,k_i}^n,\eta_{i,N}^n\right)
\label{eq21}\end{equation}
where the number of iterations is denoted by $n$. This optimization step admits a decomposition into subproblems, which can be computed in parallel for every agent $i\in\mathcal{M}$ 
\begin{equation}\begin{gathered}\left\{\boldsymbol{x}_{i,k_i},\boldsymbol{u}_{i,k_i};t_{i,N}\right\}^{n+1}=\mathrm{argmin}\sum_{k=1}^{N-1}\left[\widehat{l}_i(\boldsymbol{x}_{i,k_i},\boldsymbol{u}_{i,k_i};t_{i,N})\right]+\widehat{\phi}_i(\boldsymbol{x}_{i,N};t_{i,N})
\\
s.t.\quad\boldsymbol{x}_{i,k_{i+1}}=\boldsymbol{f}_i(\boldsymbol{x}_{i,k_i},\boldsymbol{u}_{i,k_i};t_{i,N})\label{eq22}\end{gathered}\end{equation}
where the running cost $\widehat{l}_i(\boldsymbol{x}_{i,k_i},\boldsymbol{u}_{i,k_i};t_{i,N})$ and the terminal cost $\widehat{\phi}_i(\boldsymbol{x}_{i,N};t_{i,N})$ are revised as 
\begin{equation}\begin{gathered}\hat{l}_i(\boldsymbol{x}_{i,k_i},\boldsymbol{u}_{i,k_i};t_{i,N})=l_i(\boldsymbol{x}_{i,k_i},\boldsymbol{u}_{i,k_i};t_{i,N})+\frac{\rho}{2}\left\|\boldsymbol{x}_{i,k_i}-\widetilde{\boldsymbol{x}}_{i,k_i}+\frac{\boldsymbol{\lambda}_{i,k_i}}{\rho}\right\|_2^2+\frac{\tau}{2}\left\|\boldsymbol{u}_{i,k_i}-\widetilde{\boldsymbol{u}}_{i,k_i}+\frac{\boldsymbol{\zeta}_{i,k}}{\tau}\right\|_2^2
\\
\widehat{\phi}_i(\boldsymbol{x}_{i,N};t_{i,N})=\phi_i(\boldsymbol{x}_{i,N};t_{i,N})+\frac{\rho}{2}\left\|\boldsymbol{x}_{i,N}-\widetilde{\boldsymbol{x}}_{i,N}+\frac{\boldsymbol{\lambda}_{i,N}}{\rho}\right\|_2^2+\frac{\sigma}{2}\left\|t_{i,N}-\widetilde{t}_{i,N}+\frac{\nu_{i,N}}{\sigma}\right\|_2^2\label{eq23}\end{gathered}\end{equation}

Each agent $i\in\mathcal{M}$ solves its respective subproblem in parallel via the PDDP method, whereby the derivatives of the action-value function $Q(\boldsymbol{x}_{i,k_i},\boldsymbol{u}_{i,k_i};\boldsymbol{\theta}_i)$ are modified as 
\begin{equation}\begin{aligned}
 & Q_{\boldsymbol{x}_{i},k_{i}}=l_{\boldsymbol{x}_{i},k_{i}}+\boldsymbol{f}_{\boldsymbol{x}_{i},k_{i}}^{T}V_{\boldsymbol{x}_{i},k_{i+1}}+\rho(\boldsymbol{x}_{i,k_{i}}-\widetilde{\boldsymbol{x}}_{i,k_{i}})+\boldsymbol{\lambda}_{i,k_{i}} \\
 & Q_{\boldsymbol{u}_{i},k_{i}}=l_{\boldsymbol{u}_{i},k}+\boldsymbol{f}_{\boldsymbol{u}_{i},k}^{T}V_{\boldsymbol{x}_{i},k_{i+1}}+\tau(\boldsymbol{u}_{i,k_{i}}-\widetilde{\boldsymbol{u}}_{i,k_{i}})+\boldsymbol{\zeta}_{i,k_{i}} \\
 & Q_{\boldsymbol{\theta}_i,k_i}=l_{\boldsymbol{\theta}_i,k_i}+V_{\boldsymbol{\theta}_i,k_{i+1}}+\boldsymbol{f}_{\boldsymbol{\theta}_i,k}^TV_{\boldsymbol{x}_i,k_{i+1}} \\
 & Q_{\boldsymbol{x}_{i}\boldsymbol{x}_{i},k_{i}}=l_{\boldsymbol{x}_{i}\boldsymbol{x}_{i},k_{i}}+\boldsymbol{f}_{\boldsymbol{x}_{i},k_{i}}^{T}V_{\boldsymbol{x}_{i}\boldsymbol{x}_{i},k_{i+1}}\boldsymbol{f}_{x_{i},k_{i}}+\rho\boldsymbol{I} \\
 & Q_{\boldsymbol{u}_i\boldsymbol{u}_i,k_i}=l_{\boldsymbol{u}_i\boldsymbol{u}_i,k_i}+\boldsymbol{f}_{\boldsymbol{u}_i,k_i}^TV_{\boldsymbol{x}_i\boldsymbol{x}_i,k_{i+1}}\boldsymbol{f}_{\boldsymbol{u}_i,k_i}+\tau\boldsymbol{I} \\
 & Q_{\boldsymbol{\theta}_{i}\boldsymbol{\theta}_{i,k_{i}}}=l_{\boldsymbol{\theta}_{i}\boldsymbol{\theta}_{i},k_{i}}+V_{\boldsymbol{\theta}_{i}\boldsymbol{\theta}_{i},k_{i+1}}+\boldsymbol{f}_{\boldsymbol{\theta}_{i},k_{i}}^{T}V_{\boldsymbol{x}_{i}\boldsymbol{x}_{i},k_{i+1}}\boldsymbol{f}_{\boldsymbol{\theta}_{i},k_{i}}+\boldsymbol{f}_{\boldsymbol{\theta}_{i},k_{i}}^{T}V_{\boldsymbol{x}_{i}\boldsymbol{\theta}_{i},k_{i+1}}+V_{\boldsymbol{\theta}_{i}\boldsymbol{x}_{i},k_{i+1}}\boldsymbol{f}_{\boldsymbol{\theta}_{i},k_{i}} \\
 & Q_{\boldsymbol{x}_{i}\boldsymbol{u}_{i},k_{i}}=l_{\boldsymbol{x}_{i}\boldsymbol{u}_{i},k_{i}}+\boldsymbol{f}_{\boldsymbol{x}_{i},k_{i}}^{T}V_{\boldsymbol{x}_{i}\boldsymbol{x}_{i},k_{i+1}}\boldsymbol{f}_{\boldsymbol{u}_{i},k_{i}}=Q_{\boldsymbol{u}_{i}x_{i},k_{i}}^{T} \\
 & Q_{\boldsymbol{x}_{i}\boldsymbol{\theta}_{i},k_{i}}=l_{\boldsymbol{x}_{i}\boldsymbol{\theta}_{i},k_{i}}+\boldsymbol{f}_{\boldsymbol{x}_{i},k_{i}}^{T}V_{\boldsymbol{x}_{i}\boldsymbol{x}_{i},k_{i+1}}f_{\boldsymbol{\theta}_{i},k_{i}}+\boldsymbol{f}_{\boldsymbol{x}_{i},k_{i}}^{T}V_{\boldsymbol{x}_{i}\boldsymbol{\theta}_{i},k_{i+1}}=Q_{\boldsymbol{\theta}_{i}\boldsymbol{x}_{i},k_{i}}^{T} \\
 & Q_{\boldsymbol{u}_i\boldsymbol{\theta}_i,k_i}=l_{\boldsymbol{u}_i\boldsymbol{\theta}_i,k_i}+\boldsymbol{f}_{\boldsymbol{u}_i,k_i}^TV_{\boldsymbol{x}_i\boldsymbol{x}_i,k_{i+1}}\boldsymbol{f}_{\boldsymbol{\theta}_i,k_i}+\boldsymbol{f}_{\boldsymbol{u}_i,k_i}^TV_{\boldsymbol{u}_i\boldsymbol{\theta}_i,k_{i+1}}=Q_{\boldsymbol{\theta}_i\boldsymbol{u}_i,k_i}^T
\label{eq24}\end{aligned}\end{equation}

The terminal conditions for the value function $V(\boldsymbol{x}_{i,N};t_{i,N})$ and its derivatives are specified through the following equations: 
\begin{equation}\begin{aligned}
 & V\left(\boldsymbol{x}_{i,N};t_{i,N}\right)=\phi_{i}\left(\boldsymbol{x}_{i,N};t_{i,N}\right)+\frac{\rho}{2}\Vert\boldsymbol{x}_{i,N}-\widetilde{\boldsymbol{x}}_{i,N}+\frac{\boldsymbol{\lambda}_{i,N}}{\rho}\Vert_{2}^{2}+\frac{\sigma}{2}\Vert t_{i,N}-\widetilde{t}_{i,N}+\frac{\nu_{i,N}}{\sigma}\Vert_{2}^{2} \\
 & V_{\boldsymbol{x}i,N}=\phi_{\boldsymbol{x}_i,N}+\rho\left(\boldsymbol{x}_{i,N-1}-\widetilde{\boldsymbol{x}}_{i,N-1}\right)+\boldsymbol{\lambda}_{i,N-1} \\
 & V_{\boldsymbol{\theta}_i,N}=\phi_{\boldsymbol{\theta}_i,N}+\sigma\left(t_{i,N-1}-\tilde{t}_{i,N-1}\right)+\nu_{i,N-1} \\
 & V_{x_ix_i,N}=\phi_{x_ix_i,N}+\rho\boldsymbol{I}_{p_i\times p_i} \\
 & V_{x_i\boldsymbol{\theta}_i,N}=\phi_{x_i\boldsymbol{\theta}_i,N}=V_{\boldsymbol{\theta}_i\boldsymbol{x}_i,N}^T \\
 & V_{\boldsymbol{\theta}_i\boldsymbol{\theta}_i,N}=\phi_{\boldsymbol{\theta}_i\boldsymbol{\theta}_i,N}+\sigma\boldsymbol{I}_{1\times1}
\label{eq25}\end{aligned}\end{equation}

The foregoing derivation thereby addresses the dynamic constraint illustrated in Figure \ref{fig1}. Furthermore, the PDDP solver also needs to deal with the constraints $\boldsymbol{b}_{i,k_i}(\boldsymbol{u}_{i,k_i})\leq0$ shown in Figure \ref{fig1} to safeguard against divergence caused by suboptimal initial guesses. To ensure numerical stability, the updated control input is confined to a bounded region via an element-wise projection operator: 
\begin{equation}\boldsymbol{u}_{i,k_i}=\min(\max(\boldsymbol{u}_{i,k_i}+\delta\boldsymbol{u}_{i,k_i}^*,\boldsymbol{u}_{\min}),\boldsymbol{u}_{\max})\label{eq26}\end{equation}
where $\boldsymbol{u}_{\min}$ and $\boldsymbol{u}_{\max}$ denote the lower and higher boundaries of the control effort, respectively. 

In a similar manner, the update of the final time is limited to satisfy $c_{i,N}(\tilde{t}_{i,N})\leq0$ in Figure \ref{fig1} as 
\begin{equation}t_{i,N}=\min(\max(t_{i,N}+\delta t_{i,N}^*,t_{\min}),t_{\max})\label{eq27}\end{equation}
where $t_{\min}$ and $t_{\max}$ represent the lower and higher boundaries on the final time, respectively. 

Moreover, as the workers have different speeds, the $\tilde{\boldsymbol{z}}_i$’s are in general different. In other words, as in recent distributed asynchronous optimization algorithms \citep{bib16,bib17}, some workers may be using out-of-date versions of the consensus variable. The new $\boldsymbol{x}_{i,k_{i+1}}$, together with $\boldsymbol{\lambda}_{i,k_i}$, are sent to the master. Worker $i$ then waits for the next  $\boldsymbol{z}$, $\boldsymbol{s}$ update from the master before further processing. 

\subsection{Optimization Steps 2: Single-Agent Safe Update}\label{sec3:sub2}

In this update step, the remaining local constraints are handled by the async-ADMM method, with the update governed by the following rule 
\begin{equation}\left\{\widetilde{\boldsymbol{x}}_{i,k_{i}}^{a},\widetilde{\boldsymbol{u}}_{i,k_{i}},\widetilde{t}_{i,N}^{a}\right\}^{n+1}=
\mathrm{argmin}\mathscr{L}\left(\boldsymbol{x}_{i,k_{i}}^{n+1},\boldsymbol{u}_{i,k_{i}}^{n+1},t_{i,N}^{n+1},\widetilde{\boldsymbol{x}}_{i,k_{i}}^{a},\widetilde{\boldsymbol{u}}_{i,k_{i}},\widetilde{t}_{i,N}^{a},\boldsymbol{z}_{i,k_{i}}^{a,n},s_{i,N}^{a,n},\boldsymbol{\zeta}_{i,k_{i}}^{n},\boldsymbol{\lambda}_{i,k_{i}}^{n},\nu_{i,N}^{n},\boldsymbol{y}_{i,k_{i}}^{n},\eta_{i,N}^{n}\right)\label{eq28}\end{equation}

Owing to the decoupled nature of the state and control variables across agents, Problem (\ref{eq28}) can be decomposed into $M\times(N+1)$ independent state subproblems for every agent $i$ at each time instant $k_i$. These subproblems can be solved in parallel as 
\begin{equation}\begin{gathered}
\widetilde{\boldsymbol{x}}_{i,k_i}^{a,n+1} =\mathrm{argmin}\frac{\rho}{2}\left\|\boldsymbol{x}_{i,k_i}-\widetilde{\boldsymbol{x}}_{i,k_i}+\frac{\boldsymbol{\lambda}_{i,k_i}}{\rho}\right\|_2^2+\frac{\mu}{2}\left\|\widetilde{\boldsymbol{x}}_{i,k_i}^a-\boldsymbol{z}_{i,k_i}^a+\frac{\boldsymbol{y}_{i,k_i}}{\mu}\right\|_2^2 \\
 s.t.\quad\boldsymbol{a}_{i,k_i}(\widetilde{\boldsymbol{x}}_{i,k_i})\leq0,\boldsymbol{d}_{i,k_i}^a(\widetilde{\boldsymbol{x}}_{i,k_i}^a)\leq0
\label{eq29}\end{gathered}\end{equation}
Similarly, the control effort can be decomposed into $M\times N$ smaller problems for solving 
\begin{equation}\begin{gathered}
\widetilde{\boldsymbol{u}}_{i,k_i}^{n+1}=\mathrm{argmin}\frac{\tau}{2}\left\|\boldsymbol{u}_{i,k_i}-\widetilde{\boldsymbol{u}}_{i,k_i}+\frac{\boldsymbol{\zeta}_{i,k_i}}{\tau}\right\|_2^2 \\
s.t.\quad\boldsymbol{b}_{i,k_i}(\widetilde{\boldsymbol{u}}_{i,k_i})\leq0
\label{eq30}\end{gathered}\end{equation}
and $M$ terminal time optimization subproblems 
\begin{equation}\begin{gathered}\tilde{t}_{i,N}^{\mathrm{a},n+1}=\mathrm{argmin}\frac{\sigma}{2}\left\|t_{i,N}-\tilde{t}_{i,N}+\frac{\nu_{i,N}}{\sigma}\right\|_{2}^{2}+\frac{\gamma}{2}\left\|\tilde{t}_{i,N}^{a}-s_{i,N}^{a}+\frac{\eta_{i,N}}{\gamma}\right\|_{2}^{2}
\\
s.t.\quad c_{i,N}(\tilde{t}_{i,N})\leq0,\boldsymbol{e}_{i,N}^{a}(\tilde{t}_{i,N}^{a})=0\label{eq31}\end{gathered}\end{equation}
where the control input constraints $\boldsymbol{b}_{i,k_i}(\widetilde{\boldsymbol{u}}_{i,k_i})\leq0$ are handled following the procedure outlined in Section \ref{sec3:sub1}, while the terminal time boundary constraints $c_{i,N}(\tilde{t}_{i,N})\leq0$, coupled with $\boldsymbol{e}_{i,N}^{a}(\tilde{t}_{i,N}^{a})=0$, is addressed by the standard solver.

\subsection{Optimization Steps 3: Global Update}\label{sec3:sub3}

The master waits for the workers’$\{(\boldsymbol{x}_i,\boldsymbol{u}_i,\boldsymbol{\zeta}_i,\boldsymbol{\lambda}_i,\nu_i,\boldsymbol{y}_i,\eta_i)\}$  updates before it can update $\boldsymbol{z}$, $s$. Recall that for the synchronous ADMM, this can proceed only after the $\{\boldsymbol{x}_i\}$ updates from all $N$ workers have finished. Specifically, the master only needs to wait for a minimum of $S$ updates, where $S(\geq1)$ can be much smaller than $N$. Besides, update from every worker has to be serviced by the master at least once every $\tau$ iterations, where $\tau\geq1$ is a user-defined parameter. In other words, the $(\boldsymbol{x}_i,\boldsymbol{u}_i,\boldsymbol{\zeta}_i,\boldsymbol{\lambda}_i,\nu_i,\boldsymbol{y}_i,\eta_i)$ update from every worker $i$ can at most be $\tau$ clock cycles old (according to the master’s clock). In the implementation, a counter $\tau_i$ is kept by the master for each worker $i$. When $(\boldsymbol{x}_i,\boldsymbol{u}_i,\boldsymbol{\zeta}_i,\boldsymbol{\lambda}_i,\nu_i,\boldsymbol{y}_i,\eta_i)$ from worker $i$ arrives at the master, the corresponding $\tau_i$ is reset to 1; otherwise, $\tau_i$ is incremented by 1 as the master’s clock $k$ increments.

When both the partial barrier and bounded delay conditions are met, the master can proceed with the $\boldsymbol{z}$, $s$ update. Let $\boldsymbol{\phi}^{k}$ be the set of workers whose $(\boldsymbol{x}_i,\boldsymbol{u}_i,\boldsymbol{\zeta}_i,\boldsymbol{\lambda}_i,\nu_i,\boldsymbol{y}_i,\eta_i)$ updates have arrived at the master at (master’s) iteration $k$. The global variables $\boldsymbol{z}_{i,k}$, $s_{i,N}$ are updated, i.e.,
\begin{equation}\left\{\mathbf{z}_{i,k}^a,\mathbf{s}_{i,N}^a\right\}^{n+1}=\operatorname{argmin}\mathscr{L}\left(\boldsymbol{x}_{i,k}^{n+1},\boldsymbol{u}_{i,k}^{n+1},t_{i,N}^{n+1},\widetilde{\boldsymbol{x}}_{i,k}^{a,n+1},\widetilde{\boldsymbol{u}}_{i,k}^{n+1},\widetilde{t}_{i,N}^{a,n+1},\boldsymbol{z}_{i,k}^a,s_{i,N}^a,\boldsymbol{\zeta}_{i,k}^n,\boldsymbol{\lambda}_{i,k}^n,\nu_{i,N}^n,\boldsymbol{y}_{i,k}^n,\eta_{i,N}^n\right)\label{eq32}\end{equation}

According to the constraint (\ref{eq15}), we have the following 
\begin{equation}\left\{\boldsymbol{z}_{i,k},s_{i,N}\right\}^{n+1}=\operatorname{argmin}\sum_{j\in\mathcal{P}_i}\frac{\mu}{2}\left\|\widetilde{\boldsymbol{x}}_{i,k}^j-\boldsymbol{z}_{i,k}+\frac{\boldsymbol{y}_{i,k}}{\mu}\right\|_2^2+\frac{\gamma}{2}\left\|\widetilde{\boldsymbol{t}}_{i,N}^j-\boldsymbol{s}_{i,N}+\frac{\boldsymbol{\eta}_{i,N}}{\gamma}\right\|_2^2\label{eq33}\end{equation}
which means that the global consensus variable at the agent $i$ forms an aggregate representation of the state from the perspectives of all its neighbors $j\in\mathcal{N}_i$. Consequently, the update rule for these global variables is derived as 
\begin{equation}\begin{gathered}\boldsymbol{z}_{i,k}^{n+1}=\frac{1}{|\mathcal{P}_{i}|}\sum_{j\in\mathcal{P}_{i}}\left(\widetilde{\boldsymbol{x}}_{i,k}^{j,n+1}+\frac{1}{\mu}\boldsymbol{y}_{i,k}^{j,n}\right)
\\
s_{i,N}^{n+1}=\frac{1}{|\mathcal{P}_{i}|}\sum_{j\in\mathcal{P}_{i}}\left(\widetilde{t}_{i,N}^{j,n+1}+\frac{1}{\gamma}\eta_{i,N}^{j,n}\right)\label{eq34}\end{gathered}\end{equation}
where $\{\boldsymbol{z}_{j,k}\}_{j\in\mathcal{N}_i\setminus\{i\}}$ of $\boldsymbol{z}_{i,k}^a$ is acquired from neighbors of agent $i$ via inter-agent communication, while $\widetilde{\boldsymbol{x}}_{i,k}^j$ is the most recent $\boldsymbol{x}_i$ received from worker $i$ by the master. As the dual variables $\boldsymbol{y}_{i,k}$ and $\eta_{i,N}$ asymptotically converge to 0 with increasing iterations $n$, it follows from (\ref{eq34}) that the global update progressively approaches a local average of all copy variables associated of agent $i$. 

Note that though as few as only $S$ fresh updates have arrived, the update in (\ref{eq32}) is still based on all the $\left\{\tilde{\boldsymbol{x}}_{i,k}^j\right\}_{i=1}^N$. Hence, it is possible that many of these $\widetilde{\boldsymbol{x}}_{i,k}^j$’s are out-of-date. Finally, the master’s clock $k$ is incremented by 1, and it sends the updated $\boldsymbol{z}_{i,k}^{n+1}$, $s_{i,N}^{n+1}$ back to only the workers in $\boldsymbol{\phi}^{k}$. In other words, those workers whose updates are not received in this iteration will not be aware of this $\boldsymbol{z}$, $s$ update. A side benefit is that some communication bandwidth can be saved. The whole procedure for the master is shown in Algorithm \ref{algo1}.

\begin{algorithm}
\caption{Communication-Aware Asynchronous Distributed Trajectory Optimization (CA-ADTO): Processing by the master}\label{algo1}
\begin{algorithmic}[1]
\State Initialize: $k=0$, $\tilde{\boldsymbol{x}}_{i,k}^\mathrm{a}\leftarrow 0$, $\tilde{t}_{i,N}^\mathrm{a}\leftarrow t_{i,N}$, $\boldsymbol{y}_{i,k}\leftarrow0$, $\eta_{i,N}\leftarrow0$.
\State \textbf{repeat}
\State \quad \textbf{repeat}
\State \quad \quad wait;
\State \quad \textbf{until} receive a minimum of $S$ updates from the workers \textbf{and} $\max(\tau_1,\tau_2,...,\tau_N)\leq\tau$;
\State \quad \textbf{for} worker $i\in\boldsymbol{\phi}^k$ \textbf{do}
\State \quad \quad $\tau_{i}\leftarrow1$;
\State \quad \quad $\tilde{\boldsymbol{x}}_{i,k}^j\leftarrow$ newly received $\boldsymbol{x}_i$ from worker $i$;
\State \quad \quad $\boldsymbol{y}_{i,k}^j$, $\eta_{i,N}^j\leftarrow$ newly received $\boldsymbol{y}_{i,k}$, $\eta_{i,N}$ from worker $i$;
\State \quad \textbf{end for}
\State \quad \textbf{for} worker $i\notin\boldsymbol{\phi}^k$ \textbf{do}
\State \quad \quad $\tau_{i}\leftarrow\tau_{i}+1$;
\State \quad \textbf{end for}
\State \quad update $\boldsymbol{z}_{i,k}^{n+1}$, $s_{i,N}^{n+1}$ using (\ref{eq34});
\State \quad broadcast $\boldsymbol{z}_{i,k}^{n+1}$, $s_{i,N}^{n+1}$ to all the workers in $\boldsymbol{\phi}^k$;
\State \quad $k\leftarrow k+1$;
\State \textbf{until} termination;
\State \textbf{output} $\boldsymbol{z}_{i,k}$, $s_{i,N}$.
\end{algorithmic}
\end{algorithm}

%\subsubsection{Example}\label{sec3:subsub3}

\begin{figure}[h]
\centering
\includegraphics[width=0.7\textwidth]{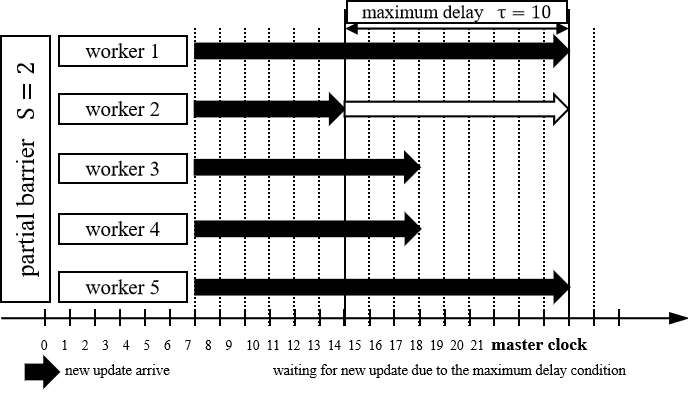}
\caption{An example showing the operation of the partial barrier and bounded delay.}\label{fig2}
\end{figure}

Figure \ref{fig2} shows an example of how the asynchronous ADMM algorithm works, with $S=2$ and $\tau=10$. When the master’s clock is at 14, updates from workers 3 and 4 arrive and the master commits an update to  $\boldsymbol{z}$. When the clock is at 21, though workers 1 and 5 have both arrived (and so meets the partial barrier condition), the $(\boldsymbol{x}_2, \boldsymbol{\lambda}_2)$ update of worker 2 has resided in the master for 10 iterations. As $\tau=10$, workers 1 and 5 have to wait until a new update from worker 2 arrives\citep{bib24}.

\subsection{Optimization Steps 4: Updating dual variable by the Worker}\label{sec3:sub4}

After receiving the updated $\boldsymbol{z}_{i,k}^{n+1}$, $s_{i,N}^{n+1}$ from the master, worker $i$ resumes its operation and updates its local copy of the dual variable:
\begin{equation}\begin{gathered}
\boldsymbol{\zeta}_{i,k_i+1}^{n+1}=\boldsymbol{\zeta}_{i,k_i}^n+\tau(\boldsymbol{u}_{i,k_i+1}^{n+1}-\tilde{\boldsymbol{u}}_{i,k}^{n+1}) \\
\boldsymbol{\lambda}_{i,k_i+1}^{n+1}=\boldsymbol{\lambda}_{i,k_i}^n+\rho(\boldsymbol{x}_{i,k_i+1}^{n+1}-\tilde{\boldsymbol{x}}_{i,k}^{n+1}) \\
\nu_{i,N}^{n+1}=\nu_{i,N}^n+\sigma(t_{i,N}^{n+1}-\tilde{t}_{i,N}^{n+1}) \\
\boldsymbol{y}_{i,k_i+1}^{n+1}=\boldsymbol{y}_{i,k_i}^n+\mu\left(\tilde{\boldsymbol{x}}_{i,k_i+1}^{a,n+1}-\boldsymbol{z}_{i,k}^{a,n+1}\right) \\
\eta_{i,N}^{n+1}=\eta_{i,N}^n+\gamma\left(\tilde{t}_{i,N}^{a,n+1}-s_{i,N}^{a,n+1}\right)
\label{eq35}\end{gathered}\end{equation}

Finally, it increments its local clock $k_i$ by 1, and update its local $\boldsymbol{x}_i$ as described in Section \ref{sec3:sub1}. The whole procedure for the worker is shown in Algorithm \ref{algo2}.

\begin{algorithm}
\caption{Communication-Aware Asynchronous Distributed Trajectory Optimization (CA-ADTO): Processing by worker $i$}\label{algo2}
\begin{algorithmic}[1]
\State $(\tilde{\boldsymbol{x}}_{i,k},\tilde{\boldsymbol{u}}_{i,k},\tilde{t}_{i,N})\leftarrow$ Solving unconstrained trajectory optimization problems for each agent $i\in\mathcal{M}$ using the DDP approach.
\State Each agent $i\in\mathcal{M}$ communicates with all its neighbors $j\in\mathcal{N}_{\mathrm{i}}\setminus\{i\}$ and receives $(\tilde{\boldsymbol{x}}_{i,k}, \tilde{t}_{i,N})$.
\State Initialize: $k=0$, $\boldsymbol{x}_{i,k}\leftarrow\tilde{\boldsymbol{x}}_{i,k}$, $\boldsymbol{u}_{i,k}\leftarrow\tilde{\boldsymbol{u}}_{i,k}$, $t_{i,N}\leftarrow\tilde{t}_{i,N}$, $\tilde{\boldsymbol{x}}_{i,k}^\mathrm{a}\leftarrow\left[\left\{\tilde{\boldsymbol{x}}_j\right\}_{j\in\mathcal{N}_i}\right]$, $\tilde{t}_{i,N}^\mathrm{a}\leftarrow\left[\left\{\tilde{t}_j\right\}_{j\in\mathcal{N}_i}\right]$, $\boldsymbol{u}_{i,k}\leftarrow\tilde{\boldsymbol{u}}_{i,k}$, $\boldsymbol{z}_{i,k}^{a}\leftarrow\tilde{\boldsymbol{x}}_{i,k}^{a}$, $\boldsymbol{\zeta}_{i,k}\leftarrow0$, $\boldsymbol{\lambda}_{i,k}\leftarrow0$, $\nu_{i,N}\leftarrow0$, $\boldsymbol{y}_{i,k}\leftarrow0$, $\eta_{i,N}\leftarrow0$.
\State \textbf{repeat}
\State \quad update $\boldsymbol{x}_{i,k_i+1}$, $\boldsymbol{u}_{i,k_i+1}$ using (\ref{eq22});
\State \quad send $\boldsymbol{y}_{k_i}$, $\eta_{i,N}$ and $\boldsymbol{x}_{i,k_i+1}$ to the master;
\State \quad \textbf{repeat}
\State \quad \quad wait;
\State \quad \textbf{until} receive $\boldsymbol{z}_{i,k}^{n+1}$, $s_{i,N}^{n+1}$ from the master;
\State \quad update $\boldsymbol{\zeta}_{i,k_i+1}^{n+1}$, $\boldsymbol{\lambda}_{i,k_i+1}^{n+1}$, $\nu_{i,N}^{n+1}$, $\boldsymbol{y}_{i,k_i+1}^{n+1}$, $\eta_{i,N}^{n+1}$ using (\ref{eq35});
\State \quad $k_i\leftarrow k_i+1$;
\State \textbf{until} termination.
\end{algorithmic}
\end{algorithm}

The convergence of the CA-ADTO algorithm is assessed by monitoring its primal and dual residuals, terminating the iterations once these values fall below a predefined tolerance. At the $n$-th iteration, these residuals are defined as follows: 
\begin{equation}\begin{aligned}
\boldsymbol{r}_{\boldsymbol{U}}^{p,n} = \boldsymbol{U}^{n} - \widetilde{\boldsymbol{U}}^{n} & \quad & \boldsymbol{r}_{\boldsymbol{U}}^{d,n} = \tau(\tilde{\boldsymbol{U}}^{n} - \tilde{\boldsymbol{U}}^{n-1}) \\
\boldsymbol{r}_{\boldsymbol{X}}^{p,n} = \boldsymbol{X}^{n} - \tilde{\boldsymbol{X}}^{n} & \quad & \boldsymbol{r}_{\boldsymbol{X}}^{d,n} = \rho(\tilde{\boldsymbol{X}}^{n} - \tilde{\boldsymbol{X}}^{n-1}) \\
\boldsymbol{r}_{\tilde{\boldsymbol{X}}}^{p,n} = \widetilde{\boldsymbol{X}}^{a,n} - \boldsymbol{Z}^{a,n} & \quad & \boldsymbol{r}_{\tilde{\boldsymbol{X}}}^{d,n} = \mu(\boldsymbol{Z}^{a,n} - \boldsymbol{Z}^{a,n-1}) \\
\boldsymbol{r}_{\boldsymbol{T}}^{p,n} = \boldsymbol{T}^{n} - \tilde{\boldsymbol{T}}^{n} & \quad & \boldsymbol{r}_{\boldsymbol{T}}^{d,n} = \sigma(\tilde{\boldsymbol{T}}^{n} - \tilde{\boldsymbol{T}}^{n-1}) \\
\boldsymbol{r}_{\tilde{\boldsymbol{T}}}^{p,n} = \tilde{\boldsymbol{T}}^{a,n} - \boldsymbol{S}^{a,n} & \quad & \boldsymbol{r}_{\tilde{\boldsymbol{T}}}^{d,n} = \gamma(\boldsymbol{S}^{n} - \boldsymbol{S}^{a,n-1})
\label{eq36}\end{aligned}
\end{equation}
where $\boldsymbol{r}_{(\cdot)}^{p,n}$ represents the primal residual and $\boldsymbol{r}_{(\cdot)}^{d,n}$ denotes the dual residual. $\boldsymbol{U}= \begin{bmatrix} \boldsymbol{U}_1^\mathrm{T},...,\boldsymbol{U}_i^\mathrm{T},...,\boldsymbol{U}_M^\mathrm{T} \end{bmatrix}^\mathrm{T}$ is the slack of the control variables from all agents and the entire time history. Similarly, we can define other vectors $\boldsymbol{X}$, $\boldsymbol{Z}$, $\boldsymbol{T}$, $\boldsymbol{S}$. Specifically, the primal residual $\boldsymbol{r}_{(\cdot)}^{p,n}$ asymptotically approaches zero upon satisfaction of the consensus constraints in Problem (\ref{eq19}), while the dual residual $\boldsymbol{r}_{(\cdot)}^{d,n}$ tends to zero as the solution converges to the optimum. The corresponding stopping criterion is therefore: 
\begin{equation}\begin{aligned} 
& \|\boldsymbol{r}_{\boldsymbol{U}}^{p,n}\|_2\leq\sqrt{\mathcal{N}_{\boldsymbol{U}}}\epsilon_{\mathrm{abs}}+\epsilon_{\mathrm{rel}}\max\{\|\boldsymbol{U}^{n}\|_2,\|\tilde{\boldsymbol{U}}^{n}\|_2\}\quad\quad\|\boldsymbol{r}_{\boldsymbol{U}}^{d,n}\|_2\leq\sqrt{\mathcal{N}_{\boldsymbol{U}}}\epsilon_{\mathrm{abs}}+\epsilon_{\mathrm{rel}}\|\Xi^{n}\|_2\\  
& \|\boldsymbol{r}_{\boldsymbol{X}}^{p,n}\|_2\leq\sqrt{\mathcal{N}_{X}}\epsilon_{\mathrm{abs}}+\epsilon_{\mathrm{rel}}\max\{\|\boldsymbol{X}^{n}\|_2,\|\tilde{\boldsymbol{X}}^{n}\|_2\}\quad\quad\|\boldsymbol{r}_{\boldsymbol{X}}^{d,n}\|_2\leq\sqrt{\mathcal{N}_{\boldsymbol{X}}}\epsilon_{\mathrm{abs}}+\epsilon_{\mathrm{rel}}\|\boldsymbol{\Lambda}^{n}\|_2\\  
& \|\boldsymbol{r}_{\boldsymbol{T}}^{p,n}\|_2\leq\sqrt{\mathcal{N}_{\boldsymbol{T}}}\epsilon_{\mathrm{abs}}+\epsilon_{\mathrm{rel}}\max\{\|\boldsymbol{T}^{n}\|_2,\|\tilde{\boldsymbol{T}}^{n}\|_2\}\quad\quad\quad\|\boldsymbol{r}_{\boldsymbol{T}}^{d,n}\|_2\leq\sqrt{\mathcal{N}_{\boldsymbol{T}}}\epsilon_{\mathrm{abs}}+\epsilon_{\mathrm{rel}}\|\boldsymbol{\Pi}^{n}\|_2\\  
& \|\boldsymbol{r}_{\tilde{\boldsymbol{X}}}^{p,n}\|_2\leq\sqrt{\mathcal{N}_{\tilde{\boldsymbol{X}}^{a}}\epsilon_{\mathrm{abs}}}+\epsilon_{\mathrm{rel}}\max\{\|\tilde{\tilde{\boldsymbol{X}}}^{a,n}\|_2,\|\boldsymbol{Z}^{a,n}\|_2\}\quad\|\boldsymbol{r}_{\tilde{\boldsymbol{X}}}^{d,n}\|_2\leq\sqrt{\mathcal{N}_{\tilde{\boldsymbol{X}}^{a}}}\epsilon_{\mathrm{abs}}+\epsilon_{\mathrm{rel}}\|\boldsymbol{Y}^{n}\|_2\\  
& \|\boldsymbol{r}_{\tilde{\boldsymbol{T}}}^{p,n}\|_2\leq\sqrt{\mathcal{N}_{\tilde{\boldsymbol{T}}^{a}}\epsilon_{\mathrm{abs}}}+\epsilon_{\mathrm{rel}}\max\{\|\tilde{\boldsymbol{T}}^{a,n}\|_2,\|\boldsymbol{S}^{a,n}\|_2\}\quad\|\boldsymbol{r}_{\tilde{\boldsymbol{T}}}^{d,n}\|_2\leq\sqrt{\mathcal{N}_{\tilde{\boldsymbol{T}}^{a}}}\epsilon_{\mathrm{abs}}+\epsilon_{\mathrm{rel}}\|\boldsymbol{V}^{n}\|_2
\label{eq37}\end{aligned}\end{equation}
where $\epsilon_{\mathrm{abs}}\geq0$ and $\epsilon_{\mathrm{rel}}\geq0$ represent the absolute and relative tolerances, respectively, while the coefficient $\mathcal{N}_{(\cdot)}$ signifies the dimension of its corresponding vector. The selection of appropriate tolerance values is application-dependent and scales with the magnitude of the typical primal and dual variables.

\section{Numerical Simulation Studies}\label{sec4}

This section rigorously validates the efficacy of the proposed algorithm through a series of numerical simulations involving a UAV swarm. We first demonstrate the capability of our CA-ADTO algorithm to handle a typical scenario featuring non-convex dynamics, path constraints, inter-agent constraints, terminal time constraints, and communication connection probability. Subsequently, we conducted a comparative study against D-PDDP, validating that the proposed algorithm effectively enhances computational efficiency. Furthermore, we investigated the influence of various parameters within CA-ADTO on its algorithmic performance.

\subsection{Mission Scenario Setting}\label{sec4:sub1}

We model each UAV in the swarm as an ideal mass point moving at a constant speed $V$ in a two-dimensional plane, neglecting its attitude dynamics. The kinematic model is governed by 
\begin{equation}\begin{aligned}
 & \dot{x}_{i}  =V\mathrm{cos}\theta_i \\
 & \dot{y}_{i} =V\mathrm{sin}\theta_i \\
 & \dot{\theta}_{i}  =\omega_i
\label{eq38}\end{aligned}\end{equation}
where $[x_i,y_i,\theta_i]^T$ is the state variable of a single UAV, $[x_i,y_i]^T$ denotes the $i$-th UAV’s coordinates and $\theta_i$ is the flight heading angle. The control input is the turning rate $\omega_i$, bounded by $|\omega_i|\leq\omega_{\max}$, where $\omega_{\max}$ is a positive constant. Additionally, we establish $t_{\min}\leq t_N\leq t_{\max}$ as the limitations for each UAV’s flying time. The terminal time $t_N$ for each UAV is constrained within $[t_{\min},t_{\max}]$.

The mission requires the swarm to reach designated terminal states while respecting various constraints. To this end, a quadratic cost function $J_i(\boldsymbol{X}_i,\boldsymbol{U}_i)$ for all UAVs in the swarm is utilized as 
\begin{equation}
J_i(\boldsymbol{X}_i,\boldsymbol{U}_i)  =\frac{1}{2}(\boldsymbol{x}_{i,N}-\boldsymbol{x}_{i,N}^d)^T\boldsymbol{W}_i^N(\boldsymbol{x}_{i,N}-\boldsymbol{x}_{i,N}^d) +\sum_{k=0}^{N-1}\left[\frac{1}{2}\boldsymbol{u}_{i,k}^T\boldsymbol{R}_i\boldsymbol{u}_{i,k}+\frac{1}{2}(\boldsymbol{x}_{i,k}-\boldsymbol{x}_i^d)^T\boldsymbol{W}_i^s(\boldsymbol{x}_{i,k}-\boldsymbol{x}_i^d)\right]
\label{eq39}\end{equation}
where $\boldsymbol{x}_{i,N}^d= \begin{bmatrix} x_i^d,y_i^d,\theta_i^d \end{bmatrix}^T$ denotes the desired terminal state vector of the $i$-th UAV. $\boldsymbol{W}_i^N$, $\boldsymbol{R}_i$ and  $\boldsymbol{W}_i^s$ are weighting diagonal matrices of the cost function. 

(1) \textit{Path constraints}: Environmental obstacles are modeled as circular no-fly zones. The collision avoidance constraint for each agent is enforced by ensuring its distance from any obstacle center remains greater than a safe threshold, i.e.,  
\begin{equation}\left(x_{i,k}-x_o\right)^2+\left(y_{i,k}-y_o\right)^2\geq (r_o+d_o)^2\label{eq40}\end{equation}
where $[x_o,y_o]^\mathrm{T}$ and $r_o$ denotes the center coordinates and radius of the obstacle $o\in O$, while $d_o$ represents the minimum safe distance that should be maintained. And the collection of obstacles in the environment are represented as $O$. 

(2) \textit{Inter-agent constraints}: Each UAV should simultaneously maintain a minimum safe distance from its neighbors to prevent collisions and remain within a maximum communication range to ensure a functional communication link, due to the physical limits of onboard devices. These dual requirements are enforced through the following constraints 
\begin{equation}\begin{gathered}\left(x_{i,k}-x_{j,k}\right)^2+\left(y_{i,k}-y_{j,k}\right)^2\geq d_{\mathrm{col}}^2
\\
\left(x_{i,k}-x_{j,k}\right)^2+\left(y_{i,k}-y_{j,k}\right)^2\leq d_{\mathrm{con}}^2\label{eq41}\end{gathered}\end{equation}
where $d_{\mathrm{col}}$ denotes the safe distance and $d_{\mathrm{con}}$ denotes maximum effective communication distance between two UAVs. 

(3) \textit{Terminal time sequence constraints}: The UAVs are required to arrive at their designated targets in a predetermined sequence, maintaining a specified time interval between arrivals. This temporal coordination constraint is enforced by the following mathematical formulation 
\begin{equation}\boldsymbol{A}_i\tilde{\boldsymbol{t}}_i^a=\boldsymbol{t}_{\delta_i}\label{eq42}\end{equation}
where $\boldsymbol{t}_{\delta_i}$ is the time interval vector and the matrix $\boldsymbol{A}_i$ represents the time series relationship between agent $i$ and its neighbors as 
\begin{equation}\mathbf{A}_i=
\begin{bmatrix}
 \\
1 & -1 & \cdots & 0 \\
 \\
1 & \vdots & \ddots & \vdots \\
 \\
\vdots & 0 & \cdots & -1 \\
 \\
1 & 0 & \cdots & 0
\end{bmatrix}\label{eq43}\end{equation}
where the $j$-th row of $\boldsymbol{A}_i$ represents the temporal relationship between UAV $i$ and its neighbor $j$, corresponding to the $j$-th row of $\boldsymbol{t}_{\delta_i}$ denoting the pre-defined time interval between $i$ and $j$. Notice that when $\boldsymbol{t}_{\delta_i}$ is a zero vector, it means that the time interval is zero and all UAVs arrive at the targets simultaneously. When $\boldsymbol{t}_{\delta_i}$ is a non-zero vector, each UAV then follows the time interval of the elements in  $\boldsymbol{t}_{\delta_i}$. To enhance algorithmic convergence in the CA-ADTO implementation, we employ a relaxation version of constraint (\ref{eq42}), expressed as 
\begin{equation}\left|\mathbf{A}_i\tilde{\boldsymbol{t}}_{i,N}^{a}-\boldsymbol{t}_{\delta_i}\right|\leq\boldsymbol{\epsilon}_{t_\delta}\label{eq44}\end{equation}
where $\boldsymbol{\epsilon}_{t_\delta}$ is a constant relaxation parameter vector. 

\subsection{Characteristics of Proposed Algorithm}\label{sec4:sub2}

In this subsection, we evaluate the distributed spatial-temporal optimization capability of the proposed CA-ADTO algorithm for UAV swarms under three different communication connection probability $p_{con}$ via numerical simulations. The simulation results of the D-PDDP algorithm, which generates trajectory plans under ideal communication conditions without delays serve as a benchmark. This comparative framework allows us to systematically analyze the effectiveness of the proposed CA-ADTO algorithm and assess the impact of communication connection probability on algorithmic performance. 

To validate the optimization capability of the CA-ADTO algorithm for time-series tasks, we design a sequential interception scenario involving six UAVs. The environment contains two circular obstacles from which all UAVs need to maintain a safe distance. The swarm is required to intercept multiple targets within the same region from different orientations, satisfying both position and flight path angle constraints. Each UAV communicates with its neighbors within a neighborhood size of $|\mathcal{N}_i|=3$, while maintaining a safe distance during the flight. The sequential nature of the interception is enforced through time-series constraints (\ref{eq44}), with all UAVs initialized at $t_{i,N}=\mathrm{9.0s}$ and required to impact targets at equal intervals of $\boldsymbol{t}_{\delta_i}=0.1\mathrm{s}$. The relaxation variable in (\ref{eq44}) is set to $\mathrm{0.01s}$ for all UAVs. Communication maintenance distance is set based on the initial position of the UAV using the closest distance criterion.

Table \ref{tab1} summarizes the specific states for each UAV, obstacle parameters (both positions and radius), the inter-agent distance requirements, and optimization parameters including flight time bounds, partial barrier, and bounded delay values. Note that the flight time (total flight duration) evolves during optimization iterations, distinct from the initial guess which remains fixed throughout the process. This configuration enables comprehensive evaluation of the algorithm's capability to handle complex spatio-temporal constraints while maintaining coordination under unreliable links. 

\begin{table}[h]
\caption{Initial conditions of the aerodynamics.}\label{tab1}%
\begin{tabular}{@{}ll@{}}
\toprule
Parameters & Value\\
\midrule
Terminal weighting matrix $\boldsymbol{W}_i^N$    & $\mathrm{diag}(25\mathbf{I}_3)$   \\
Control weighting matrix $\boldsymbol{R}_i$    & $1$   \\
State weighting matrix $\boldsymbol{W}_i^s$    & $0$   \\
Penalty $\tau^0$ of $\boldsymbol{U}_i$    & $0.2$   \\
Penalty $\rho^0$ of $\boldsymbol{X}_i$    & $2$   \\
Penalty $\mu^0$ of $\tilde{\boldsymbol{X}}_i^\mathrm{a}$    & $1$   \\
Penalty $\sigma^0$ of $\boldsymbol{T}_i$    & $2$   \\
Penalty $\gamma^0$ of $\tilde{\boldsymbol{T}}_i^\mathrm{a}$    & $1$   \\
Damping coefficient of PDDP     & $0.4$   \\
Stopping threshold $\epsilon_{\mathrm{rel}}$,$\epsilon_{\mathrm{abs}}$   & $6\times10^{-2}$,$5\times10^{-4}$   \\
Initial state of UAV $1$ $\left[x_{1_0},y_{1_0},\theta_{1_0}\right]^\mathrm{T}$    & $[\mathrm{300m},\mathrm{25m},60^\circ]^\mathrm{T}$   \\
Initial state of UAV $2$ $\left[x_{2_0},y_{2_0},\theta_{2_0}\right]^\mathrm{T}$    & $[\mathrm{437.5m},\mathrm{61.84m},120^\circ]^\mathrm{T}$   \\
Initial state of UAV $3$ $\left[x_{3_0},y_{3_0},\theta_{3_0}\right]^\mathrm{T}$    & $[\mathrm{538.16m},\mathrm{162.5m},120^\circ]^\mathrm{T}$   \\
Initial state of UAV $4$ $\left[x_{4_0},y_{4_0},\theta_{4_0}\right]^\mathrm{T}$    & $[\mathrm{575m},\mathrm{300m},180^\circ]^\mathrm{T}$   \\
Initial state of UAV $5$ $\left[x_{5_0},y_{5_0},\theta_{5_0}\right]^\mathrm{T}$    & $[\mathrm{538.16m},\mathrm{437.5m},240^\circ]^\mathrm{T}$   \\
Initial state of UAV $6$ $\left[x_{6_0},y_{6_0},\theta_{6_0}\right]^\mathrm{T}$    & $[\mathrm{437.5m},\mathrm{538.16m},270^\circ]^\mathrm{T}$   \\
Target state of UAV $1$ $\left[x_{1_f},y_{1_f},\theta_{1_f}\right]^\mathrm{T}$    & $[\mathrm{300m},\mathrm{275m},120^\circ]^\mathrm{T}$   \\
Target state of UAV $2$ $\left[x_{2_f},y_{2_f},\theta_{2_f}\right]^\mathrm{T}$    & $[\mathrm{312.5m},\mathrm{278.35m},180^\circ]^\mathrm{T}$   \\
Target state of UAV $3$ $\left[x_{3_f},y_{3_f},\theta_{3_f}\right]^\mathrm{T}$    & $[\mathrm{321.65m},\mathrm{287.5m},210^\circ]^\mathrm{T}$   \\
Target state of UAV $4$ $\left[x_{4_f},y_{4_f},\theta_{4_f}\right]^\mathrm{T}$    & $[\mathrm{325m},\mathrm{300m},240^\circ]^\mathrm{T}$   \\
Target state of UAV $5$ $\left[x_{5_f},y_{5_f},\theta_{5_f}\right]^\mathrm{T}$    & $[\mathrm{321.65m},\mathrm{312.5m},270^\circ]^\mathrm{T}$   \\
Target state of UAV $6$ $\left[x_{6_f},y_{6_f},\theta_{6_f}\right]^\mathrm{T}$    & $[\mathrm{312.5m},\mathrm{321.65m},300^\circ]^\mathrm{T}$   \\
Velocity of each UAV $V_i$    & $\mathrm{30m/s}$   \\
UAV turning maneuverability $\omega_{\max}$    & $\mathrm{0.5678rad/s}$   \\
Flight time of each UAV $t_{i,N}$    & $\mathrm{9.0s}$   \\
Lower and upper bound of flight time $[t_{\min},t_{\max}]$     & $[\mathrm{0.1s},\mathrm{20s}]$   \\
Safe distance to obstacle $d_o$ and neighbors $d_{col}$   & $\mathrm{10m}$,$\mathrm{10m}$   \\
Communication maintenance distance with neighbors $d_{con}$   & $\mathrm{390m}$   \\
Center of circular obstacle $\boldsymbol{p}_{o_1}$   & $[\mathrm{450.0m},\mathrm{400.0m}]^{\mathrm{T}}$   \\
Center of circular obstacle $\boldsymbol{p}_{o_2}$    & $[\mathrm{420.0m},\mathrm{220.0m}]^{\mathrm{T}}$   \\
Radius of circular obstacle $r_o$    & $\mathrm{40.0m}$   \\
partial barrier $S$    & $2$   \\
bounded delay $\tau$    & $10$   \\
\botrule
\end{tabular}
\end{table}

\begin{figure}[htbp]
    \centering
    \begin{subfigure}[b]{0.31\textwidth}
        \centering
        \includegraphics[height=4cm]{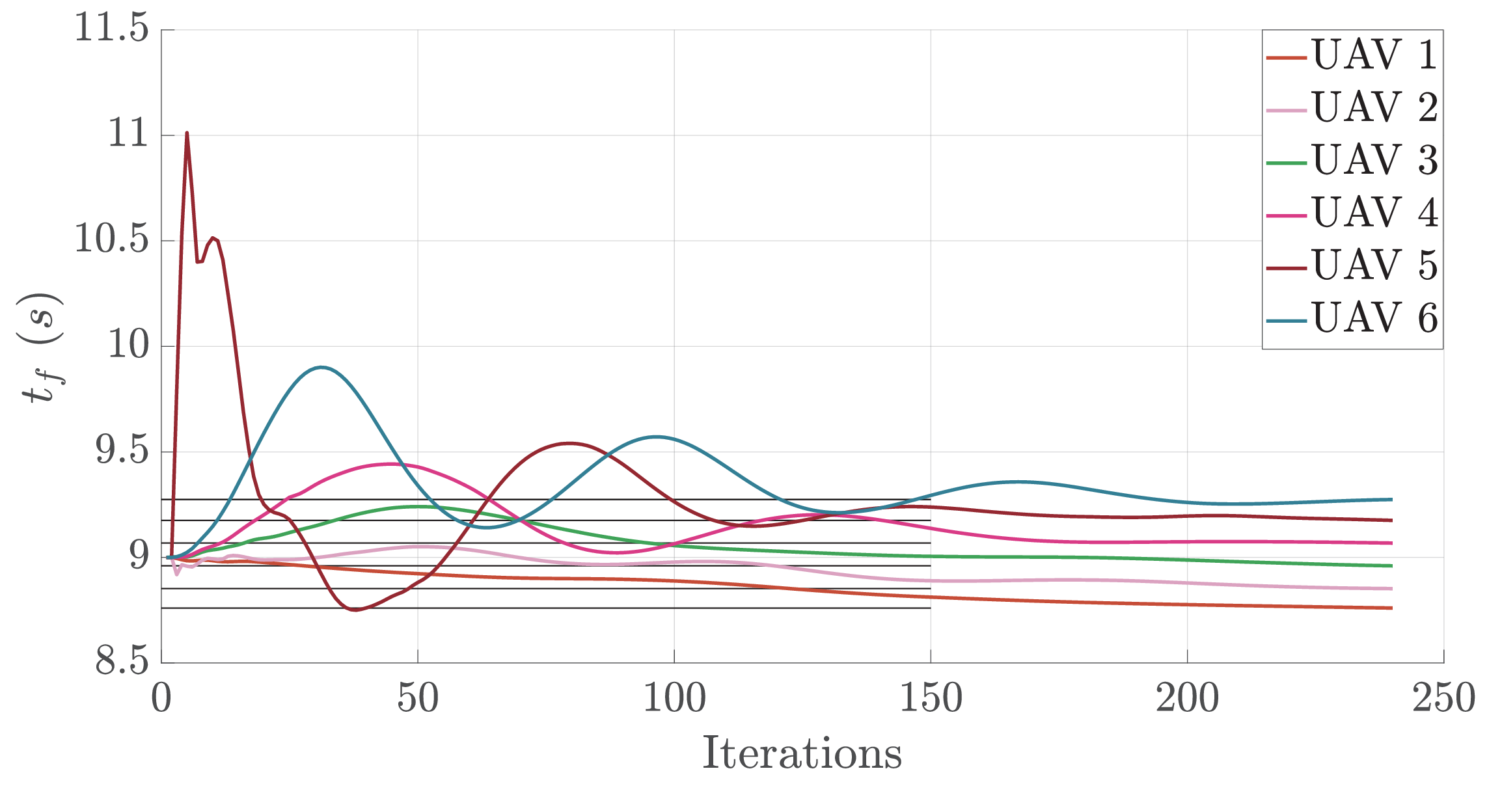}
        \caption{Convergence of the final time.}
        \label{fig3:sub1}
    \end{subfigure}
%    \hfill
    \begin{subfigure}[b]{0.31\textwidth}
        \centering
        \includegraphics[height=4cm]{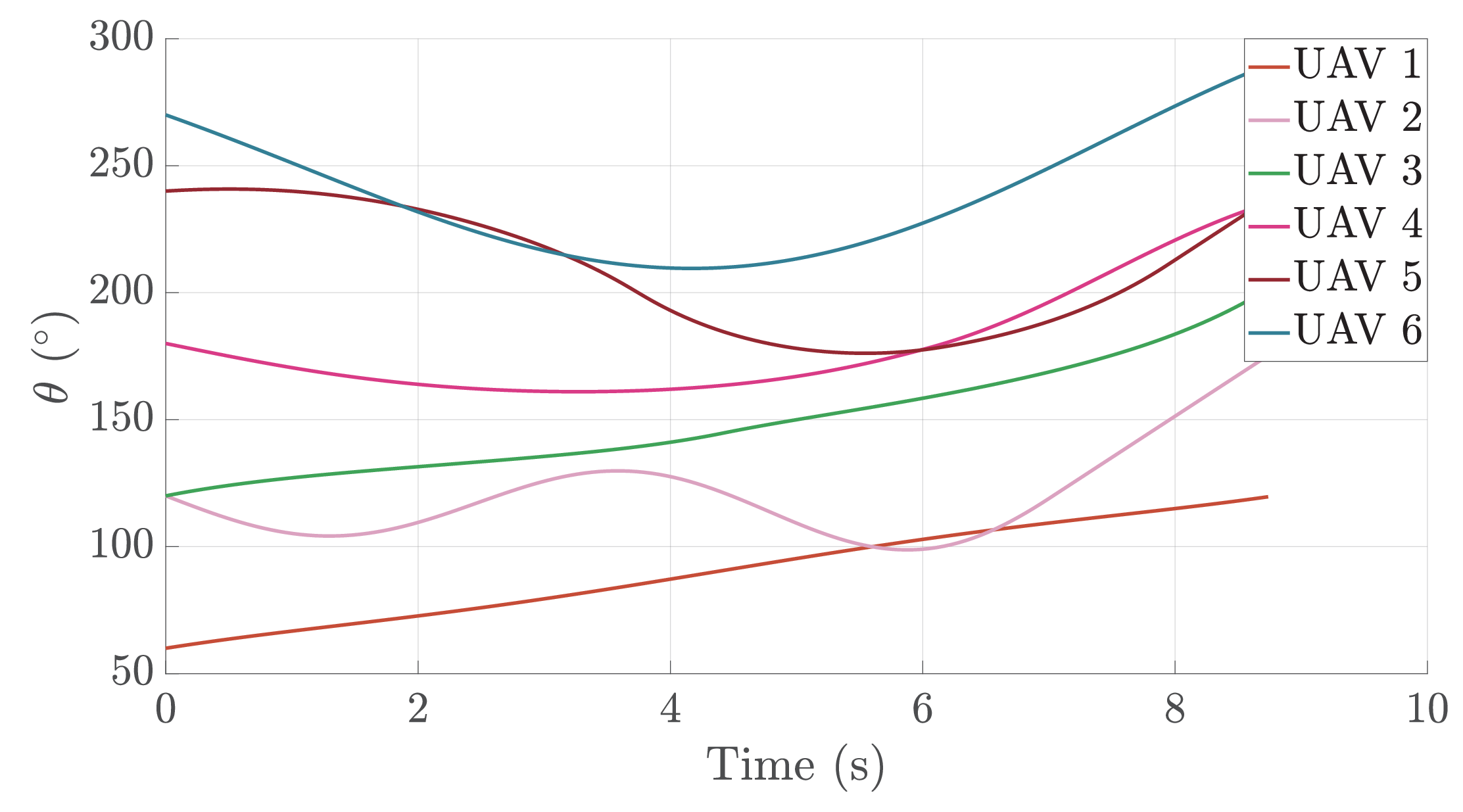}
        \caption{Time history of angular rate.}
        \label{fig3:sub2}
    \end{subfigure}
%    \vspace{0.3cm}
    \begin{subfigure}[b]{0.31\textwidth}
        \centering
        \includegraphics[height=4cm]{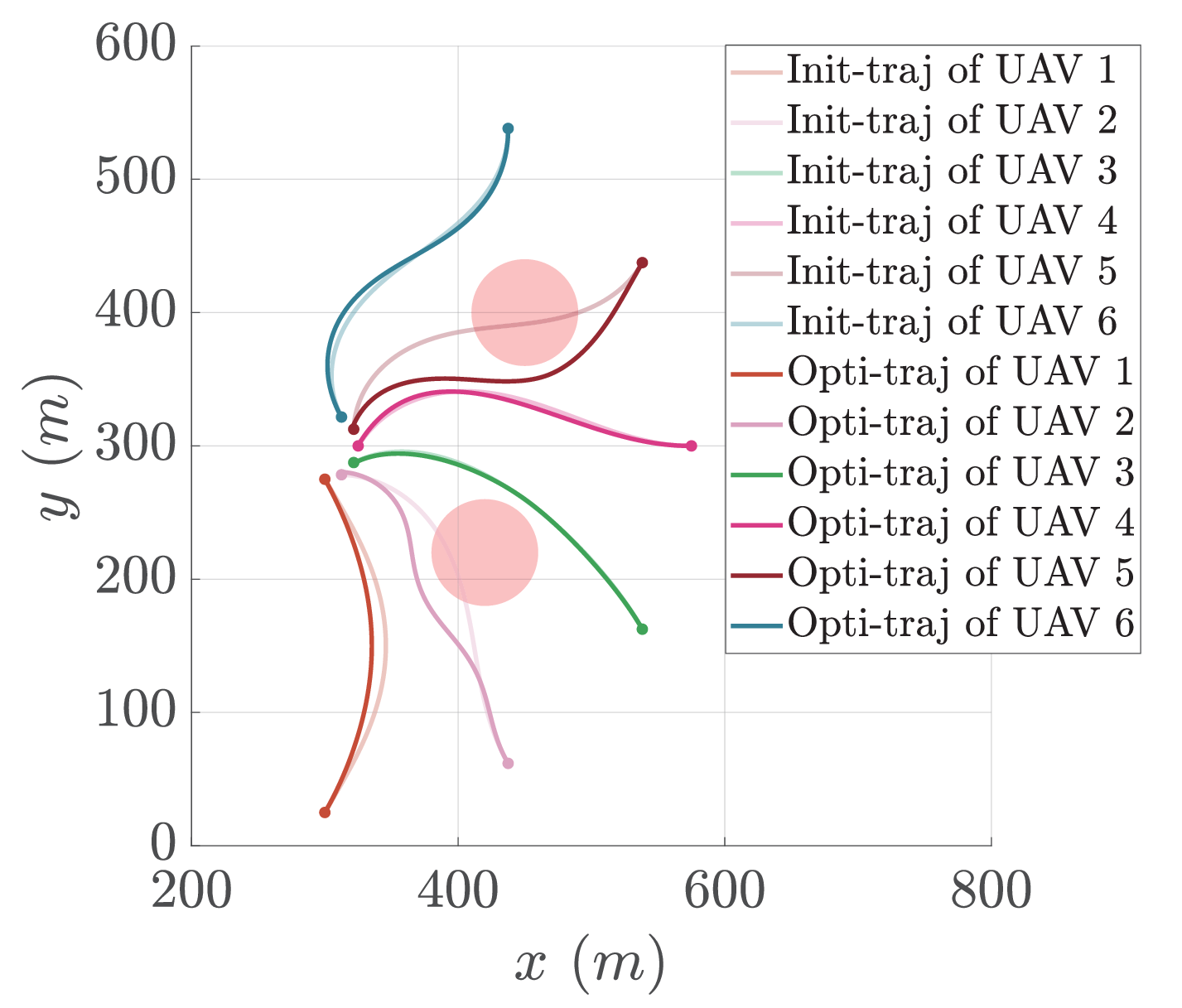}
        \caption{Trajectory.}
        \label{fig3:sub3}
    \end{subfigure}
    \caption{Distributed spatial-temporal joint optimization results ($p_{con}=\mathrm{70\%}$).}
    \label{fig3}
\end{figure}

We first conduct experiment under a communication connection probability of $p_{con}=\mathrm{70\%}$. The optimization results of the UAV swarm in the scenario are shown in Figure \ref{fig3}. It can be seen from Figure \ref{fig3:sub1} that the flight times of the five UAVs are optimized from the same initial guess $\mathrm{9.0s}$ to $\mathrm{8.76s}$, $\mathrm{8.85s}$, $\mathrm{8.96s}$, $\mathrm{9.07s}$, $\mathrm{9.17s}$, $\mathrm{9.27s}$, respectively, which satisfy the consecutive interception constraint with time interval being 0.1s between two UAVs. The control history of the six UAVs is shown in Figure \ref{fig3:sub2}. In Figure \ref{fig3:sub3}, the semi-transparent lines are the initial guesses of the flight trajectories and the opaque lines are the optimized results, from which we can see that the proposed CA-ADTO algorithm can optimize from the initialized trajectory that violates the obstacle constraints to obstacle avoidance. The trajectory results show that the trajectories of UAV $2$ and UAV $5$ change from constraint-violating trajectories to obstacle-avoiding trajectories, and the trajectory of UAV $6$ is more curved compared to the initial guess to satisfy the time sequence constraints. In summary, the results demonstrate that the proposed CA-ADTO algorithm is capable of accomplishing the UAV swarm trajectory optimization task with time sequence constraints through distributed spatial-temporal joint optimization.

\begin{figure}[htbp]
    \centering
    \begin{subfigure}[b]{0.31\textwidth}
        \centering
        \includegraphics[height=4cm]{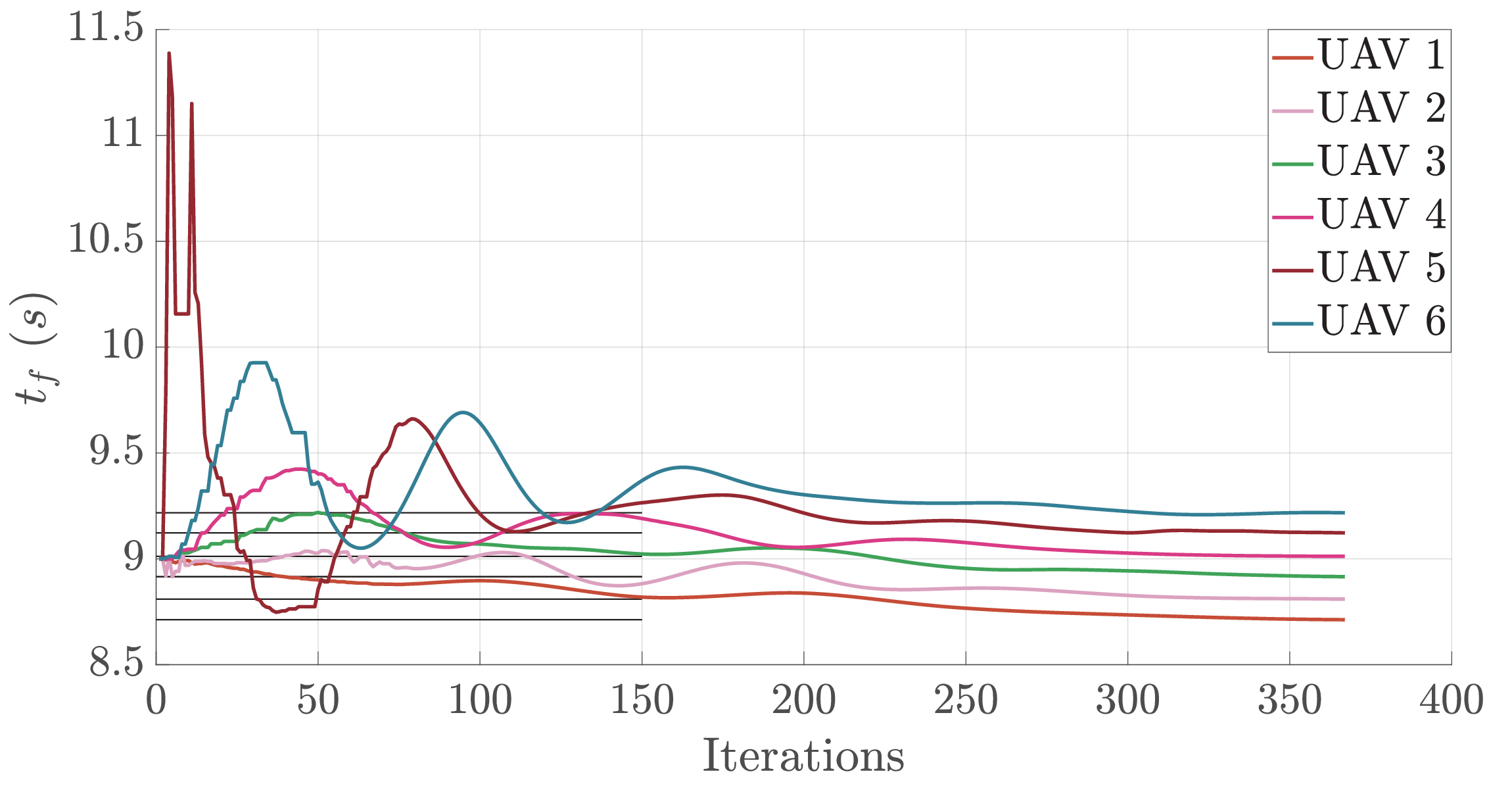}
        \caption{Convergence of the final time.}
        \label{fig4:sub1}
    \end{subfigure}
%    \hfill
    \begin{subfigure}[b]{0.31\textwidth}
        \centering
        \includegraphics[height=4cm]{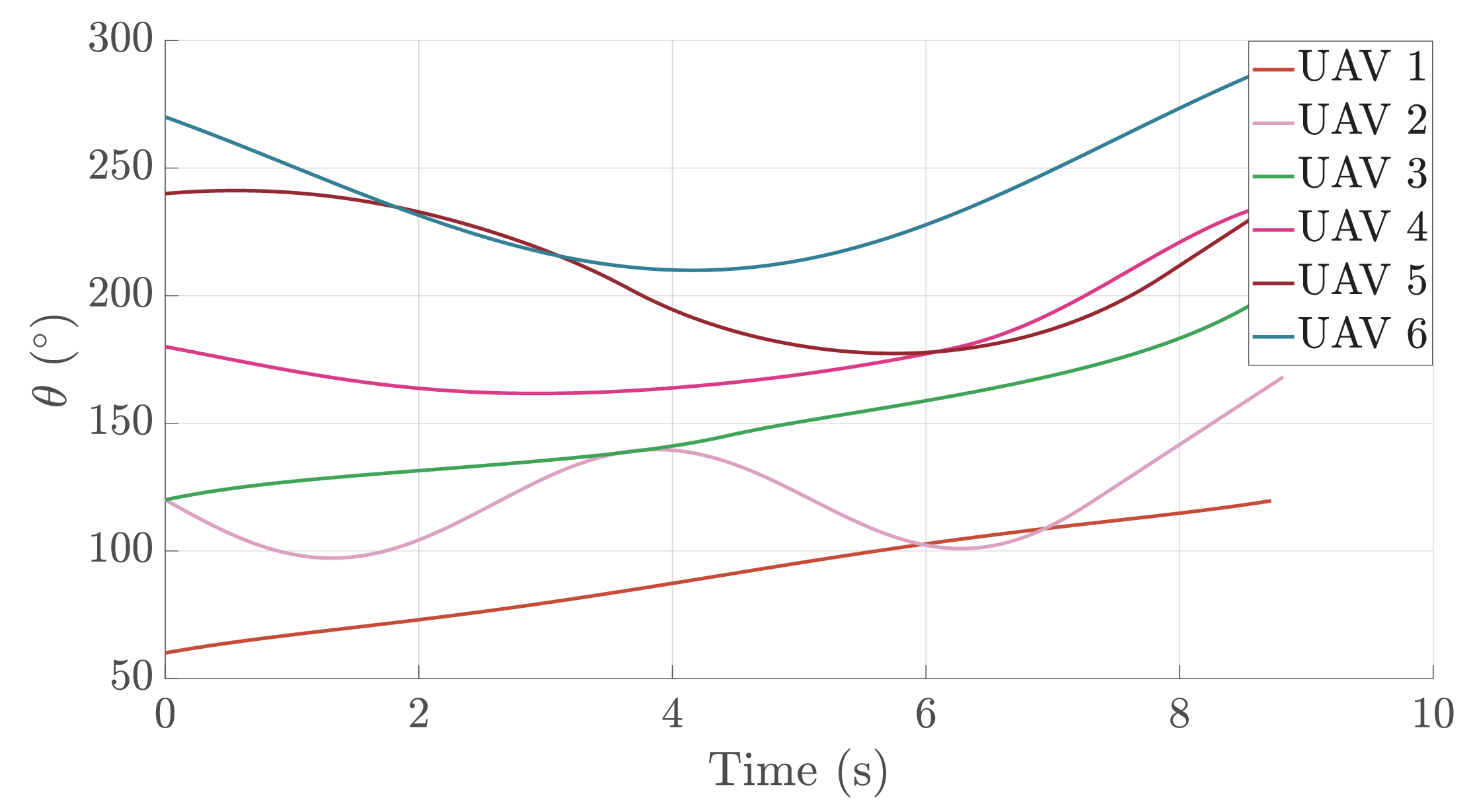}
        \caption{Time history of angular rate.}
        \label{fig4:sub2}
    \end{subfigure}
%    \vspace{0.3cm}
    \begin{subfigure}[b]{0.31\textwidth}
        \centering
        \includegraphics[height=4cm]{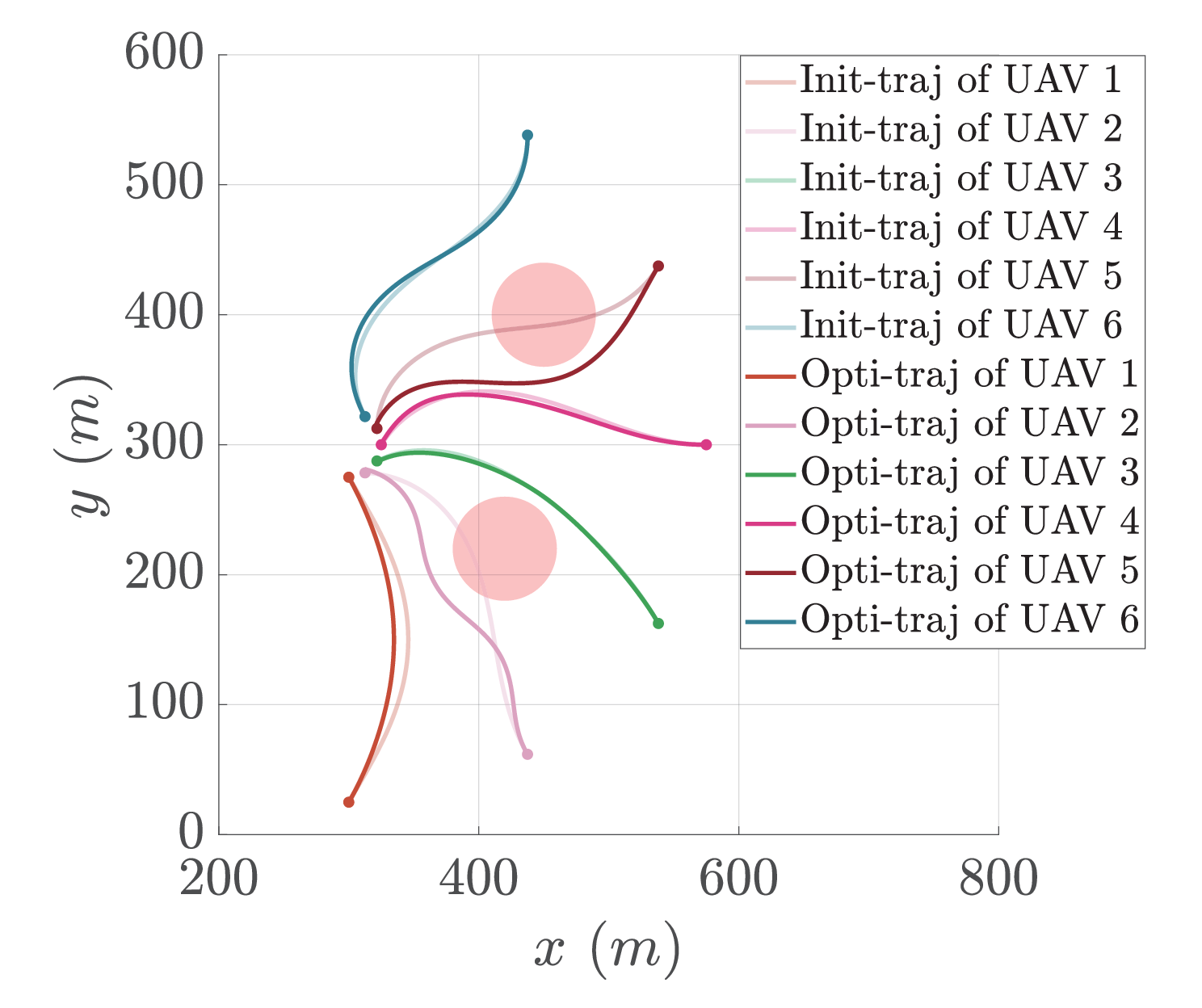}
        \caption{Trajectory.}
        \label{fig4:sub3}
    \end{subfigure}
    \caption{Distributed spatial-temporal joint optimization results ($p_{con}=\mathrm{50\%}$).}
    \label{fig4}
\end{figure}

\begin{figure}[htbp]
    \centering
    \begin{subfigure}[b]{0.31\textwidth}
        \centering
        \includegraphics[height=4cm]{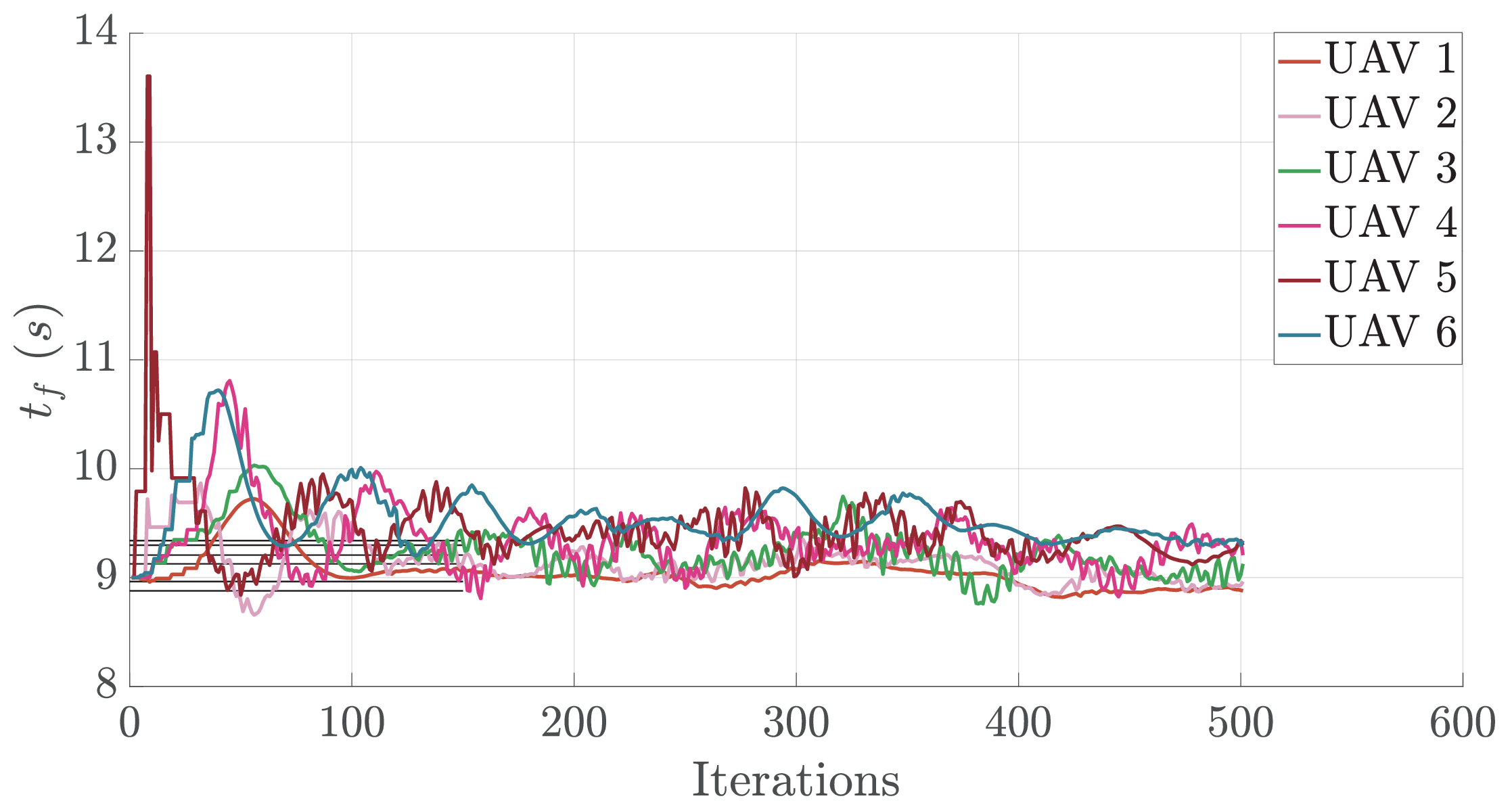}
        \caption{Convergence of the final time.}
        \label{fig5:sub1}
    \end{subfigure}
%    \hfill
    \begin{subfigure}[b]{0.31\textwidth}
        \centering
        \includegraphics[height=4cm]{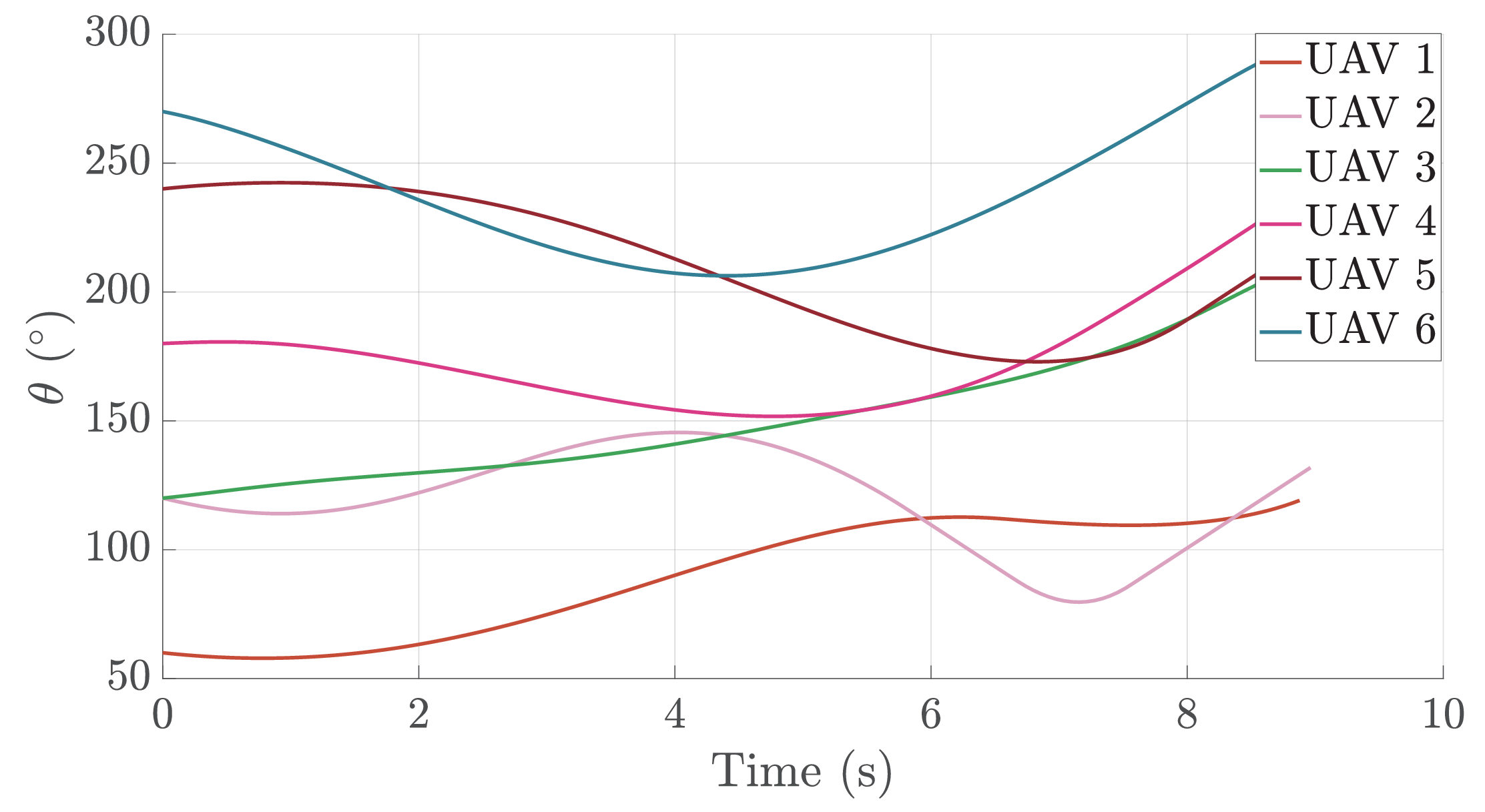}
        \caption{Time history of angular rate.}
        \label{fig5:sub2}
    \end{subfigure}
%    \vspace{0.3cm}
    \begin{subfigure}[b]{0.31\textwidth}
        \centering
        \includegraphics[height=4cm]{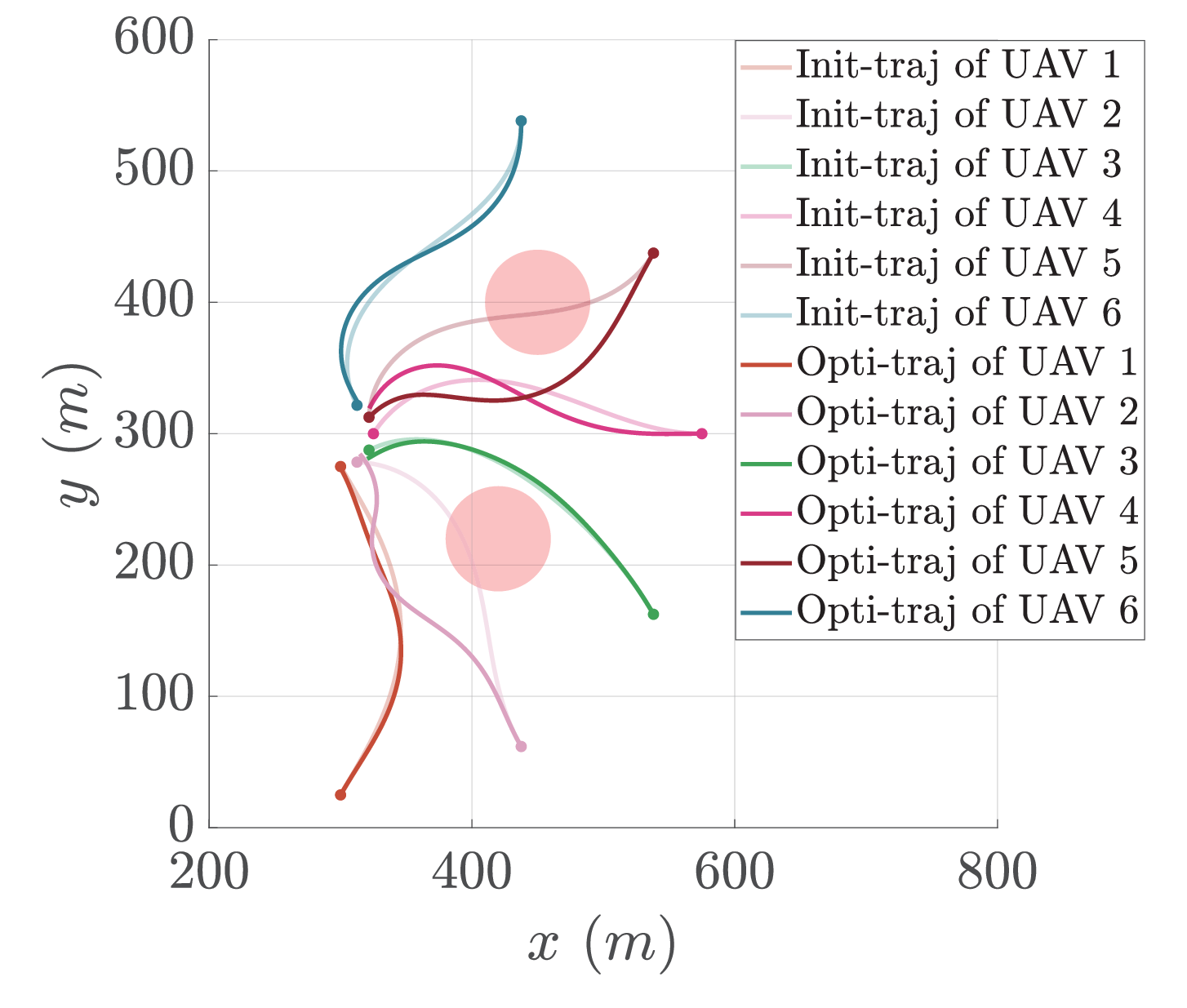}
        \caption{Trajectory.}
        \label{fig5:sub3}
    \end{subfigure}
    \caption{Distributed spatial-temporal joint optimization results ($p_{con}=\mathrm{30\%}$).}
    \label{fig5}
\end{figure}

\begin{figure}[htbp]
    \centering
    \begin{subfigure}[b]{0.31\textwidth}
        \centering
        \includegraphics[height=4cm]{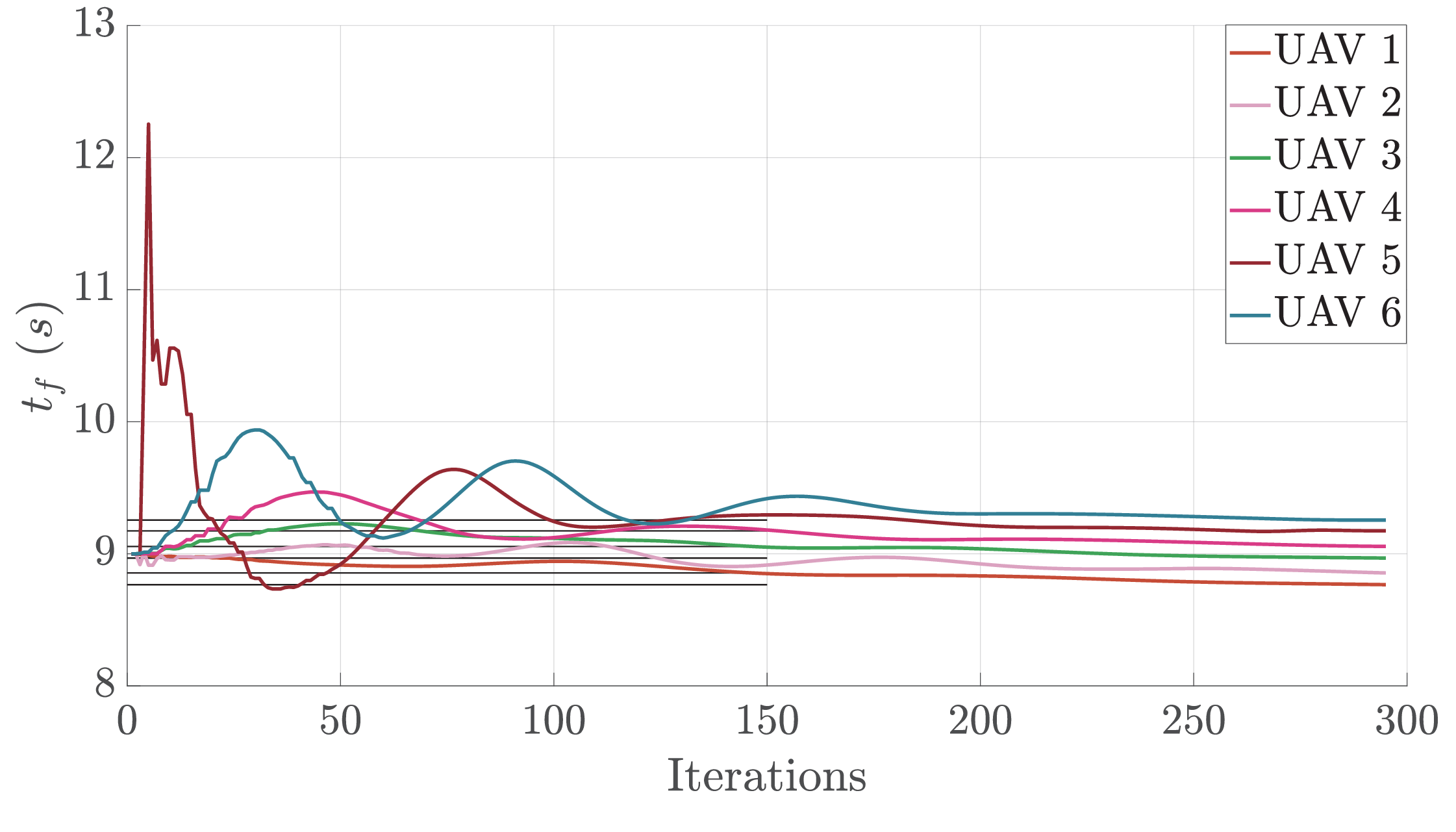}
        \caption{Convergence of the final time.}
        \label{fig6:sub1}
    \end{subfigure}
%    \hfill
    \begin{subfigure}[b]{0.31\textwidth}
        \centering
        \includegraphics[height=4cm]{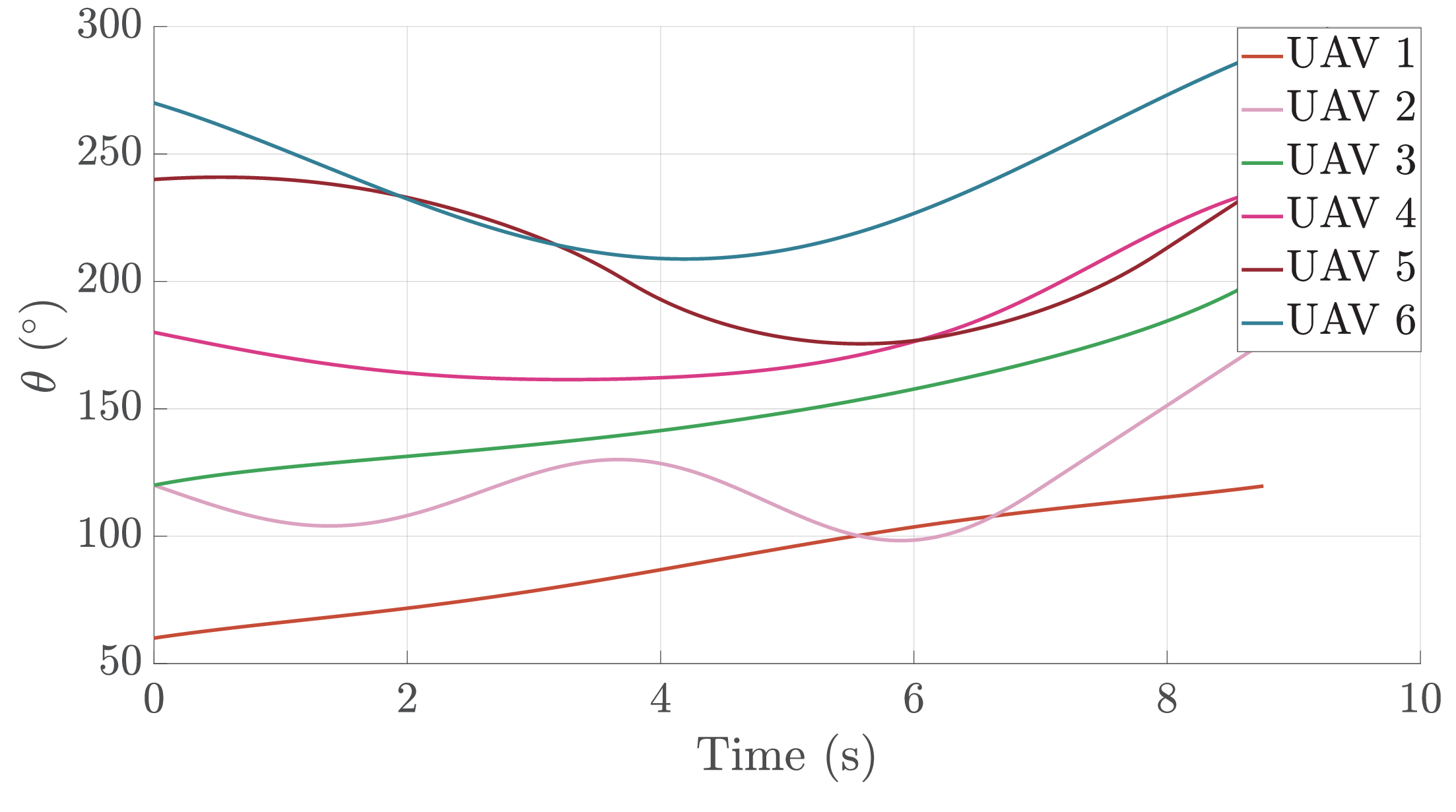}
        \caption{Time history of angular rate.}
        \label{fig6:sub2}
    \end{subfigure}
 %   \vspace{0.3cm}
    \begin{subfigure}[b]{0.31\textwidth}
        \centering
        \includegraphics[height=4cm]{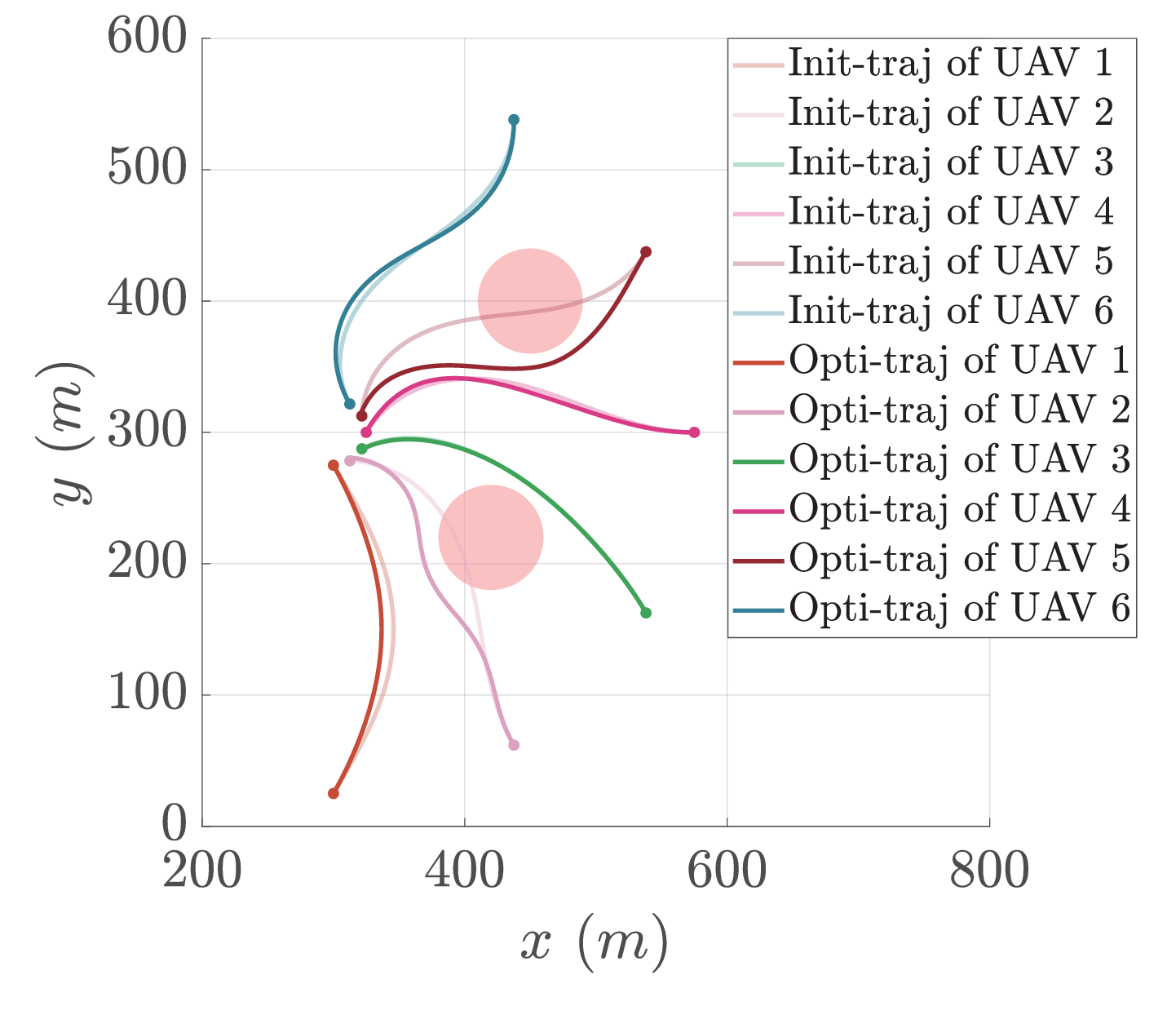}
        \caption{Trajectory.}
        \label{fig6:sub3}
    \end{subfigure}
    \caption{Distributed spatial-temporal joint optimization results (D-PDDP).}
    \label{fig6}
\end{figure}

To evaluate the impact of communication connection probability on algorithmic performance, we subsequently examine the scenario with communication connection probabilities reduced to $p_{con}=\mathrm{50\%}$ and $p_{con}=\mathrm{30\%}$. Meanwhile, to validate the advantages of our proposed algorithm, we compare the optimization results of CA-ADTO under different connection probabilities against those obtained by D-PDDP, which operates under ideal communication conditions without accounting for unreliable links. The optimization results of the UAV swarm in the scenario are shown in Figure \ref{fig4}, Figure \ref{fig5} and Figure \ref{fig6}. Figure \ref{fig4} displays the results of the CA-ADTO algorithm with a communication connection probability of $\mathrm{50\%}$ and Figure \ref{fig5} illustrates the results under a communication connection probability of $\mathrm{30\%}$. From Figure \ref{fig6}, we can see the optimization results of the D-PDDP without considering unreliable links. Together with the optimization results of the CA-ADTO algorithm at $\mathrm{70\%}$ communication connection probability from Figure \ref{fig3}, the comparative data for the four different test conditions in the scenario are summarized in Table \ref{tab2} (taking the optimization results of UAV $5$ as an example).

\begin{table}[h]
\caption{Optimization results for the scenario.}\label{tab2}%
\begin{tabular}{@{}lll@{}}
\toprule
Algorithms & Flight times & Optimization times\\
\midrule
CA-ADTO
($p_{con}=\mathrm{70\%}$)    & $\mathrm{9.17s}$   & $\mathrm{206.41s}$  \\
CA-ADTO
($p_{con}=\mathrm{50\%}$)    & $\mathrm{9.17s}$   & $\mathrm{290.18s}$  \\
CA-ADTO
($p_{con}=\mathrm{30\%}$)    & $\mathrm{9.29s}$   & $\mathrm{400.01s}$  \\
D-PDDP    & $\mathrm{9.17s}$   & $\mathrm{236.47s}$  \\
\botrule
\end{tabular}
\end{table}

In summary, the impact of the communication connection probability on the optimization performance of the CA-ADTO algorithm can be concluded as follows. When the communication connection probability is $\mathrm{70\%}$, the algorithm achieves optimization results nearly identical to those of D-PDDP without considering communication delays, while simultaneously reducing the optimization time and thereby improving overall efficiency. At a communication connection probability of $\mathrm{50\%}$, the optimization results remain largely consistent with those of D-PDDP without communication delays; however, the required trajectory optimization time increases. This configuration can reduce communication overhead to some extent and enhance communication efficiency. When the communication connection probability drops to $\mathrm{30\%}$ or below, the optimization performance deteriorates and fails to yield satisfactory results. Thus, such low communication connection probability conditions are not recommended for use.

\subsection{Characterization of CA-ADTO Parameters}\label{sec4:sub3}

In this section, the partial barrier and bounded delay parameters within the proposed algorithm are treated as adjustable variables to further investigate their influence on the performance characteristics of CA-ADTO. To maintain experimental control, the communication connection probability remains fixed at $\mathrm{70\%}$(a value established in Section \ref{sec4:sub2} as yielding favorable operational outcomes), thereby enabling isolated analysis of the parameter effects under consistent communication conditions.

\begin{figure}[htbp]
    \centering
    \begin{subfigure}[b]{0.31\textwidth}
        \centering
        \includegraphics[height=4cm]{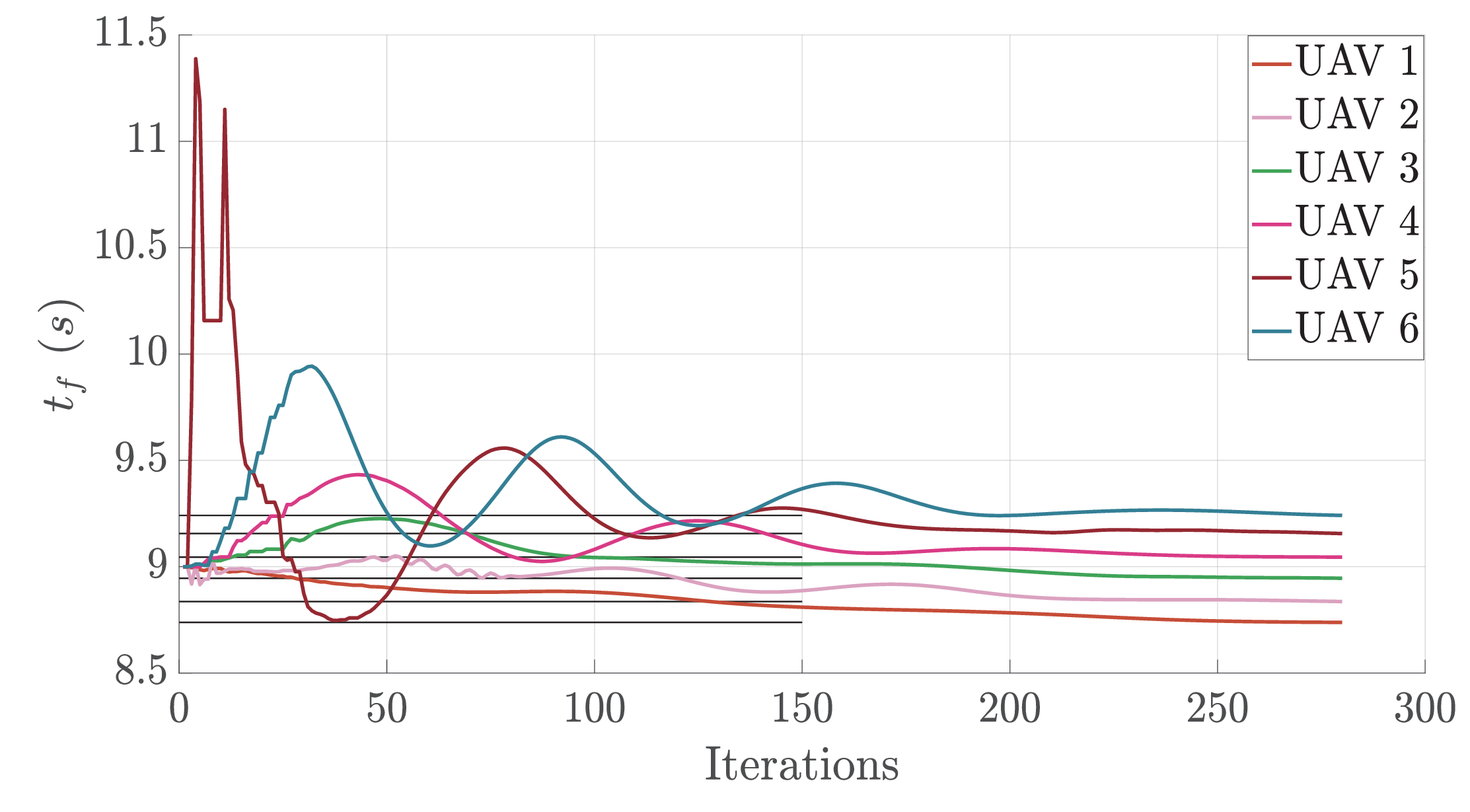}
        \caption{Convergence of the final time.}
        \label{fig7:sub1}
    \end{subfigure}
%    \hfill
    \begin{subfigure}[b]{0.31\textwidth}
        \centering
        \includegraphics[height=4cm]{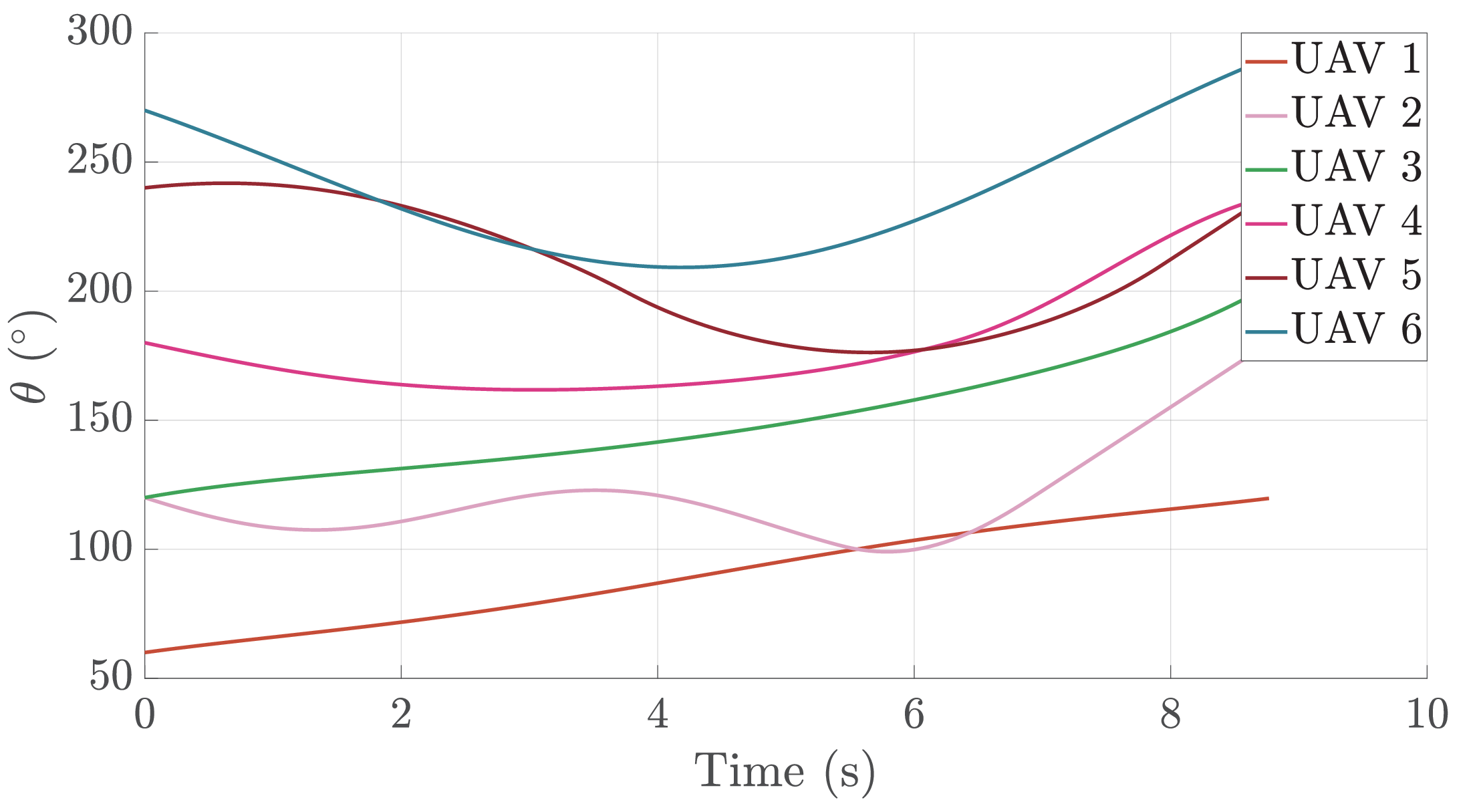}
        \caption{Time history of angular rate.}
        \label{fig7:sub2}
    \end{subfigure}
 %   \vspace{0.3cm}
    \begin{subfigure}[b]{0.31\textwidth}
        \centering
        \includegraphics[height=4cm]{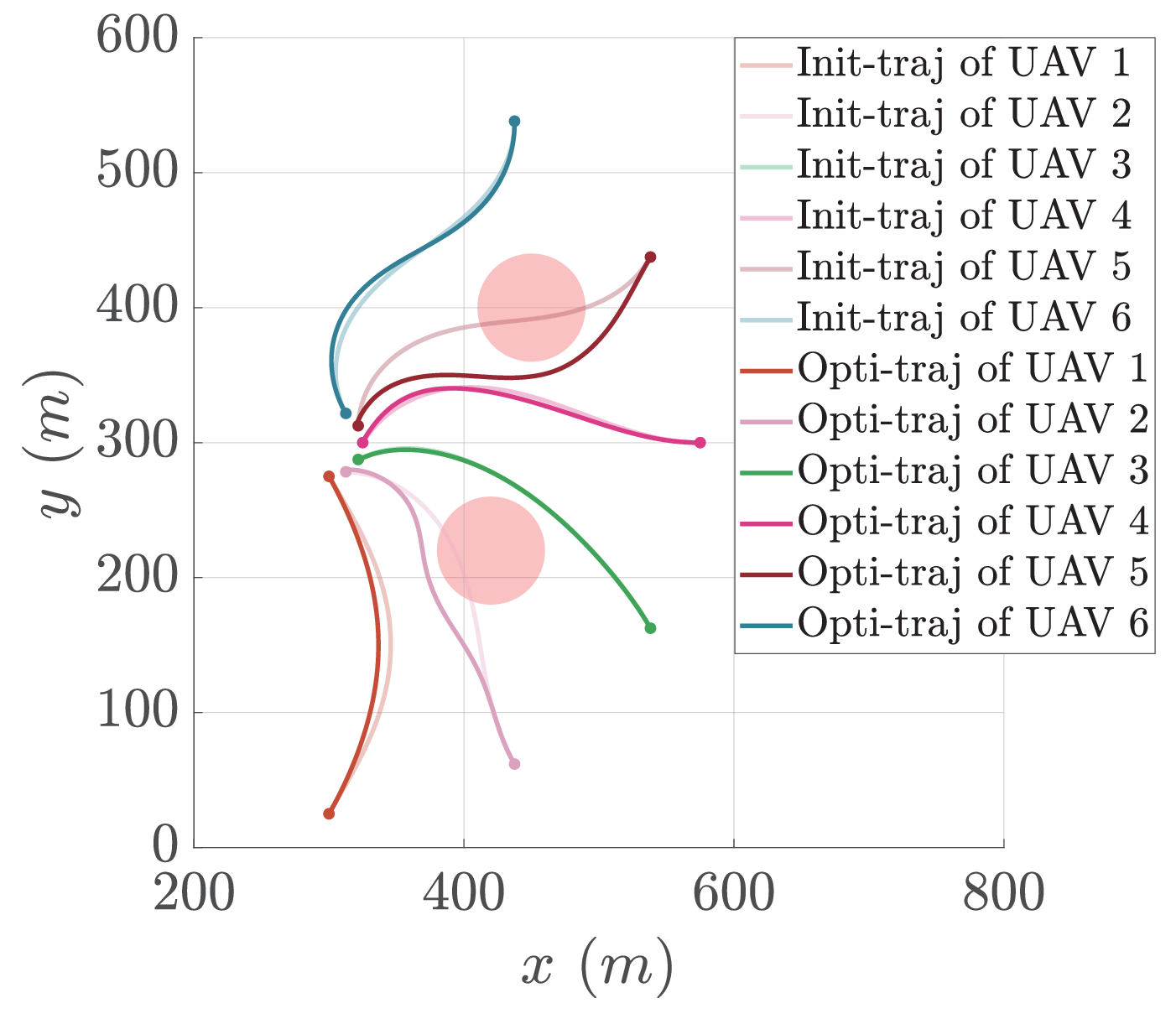}
        \caption{Trajectory.}
        \label{fig7:sub3}
    \end{subfigure}
    \caption{Distributed spatial-temporal joint optimization results
($S=N/2$, $\tau=10$).}
    \label{fig7}
\end{figure}

\begin{figure}[htbp]
    \centering
    \begin{subfigure}[b]{0.31\textwidth}
        \centering
        \includegraphics[height=4cm]{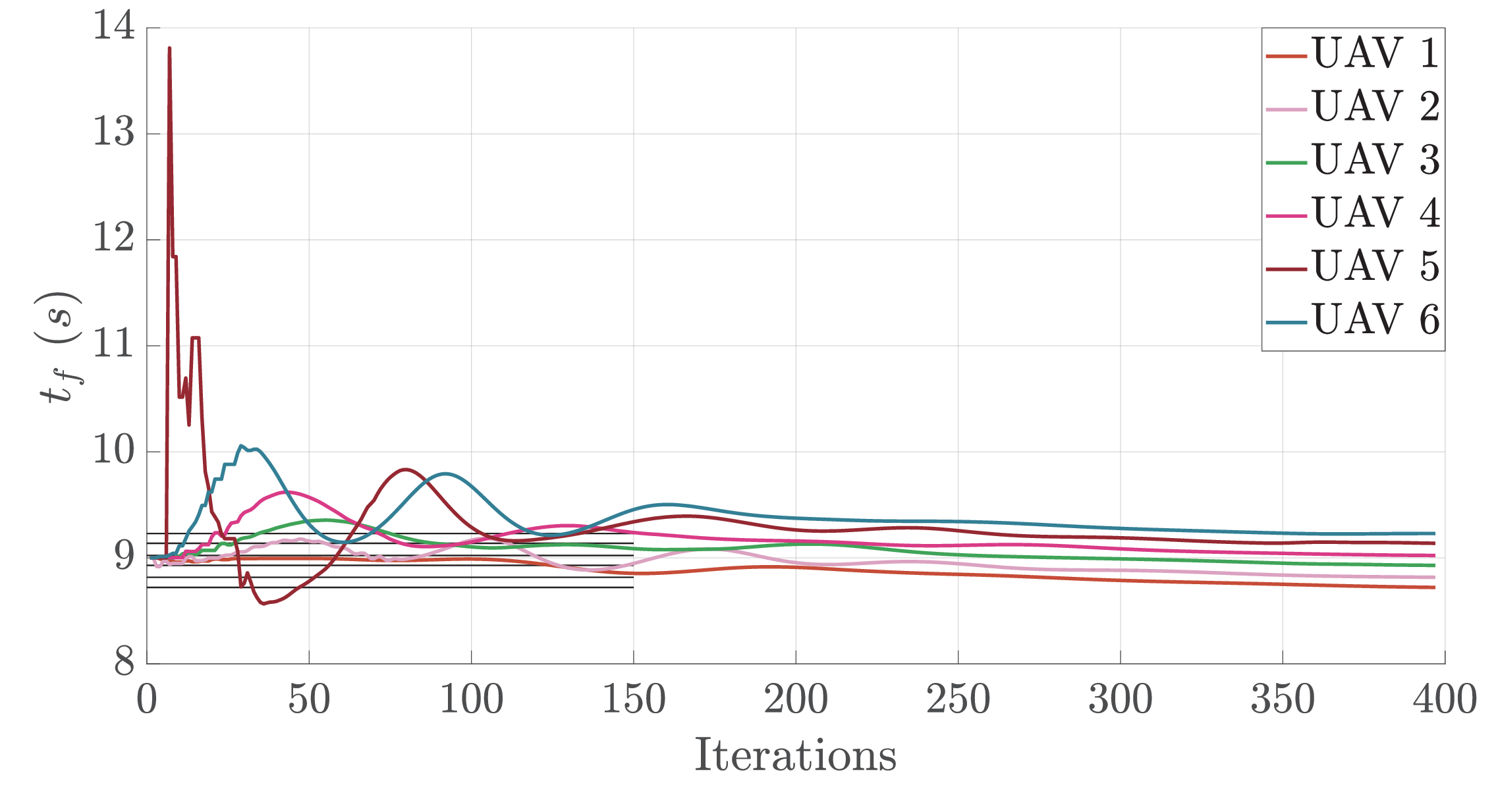}
        \caption{Convergence of the final time.}
        \label{fig8:sub1}
    \end{subfigure}
 %   \hfill
    \begin{subfigure}[b]{0.31\textwidth}
        \centering
        \includegraphics[height=4cm]{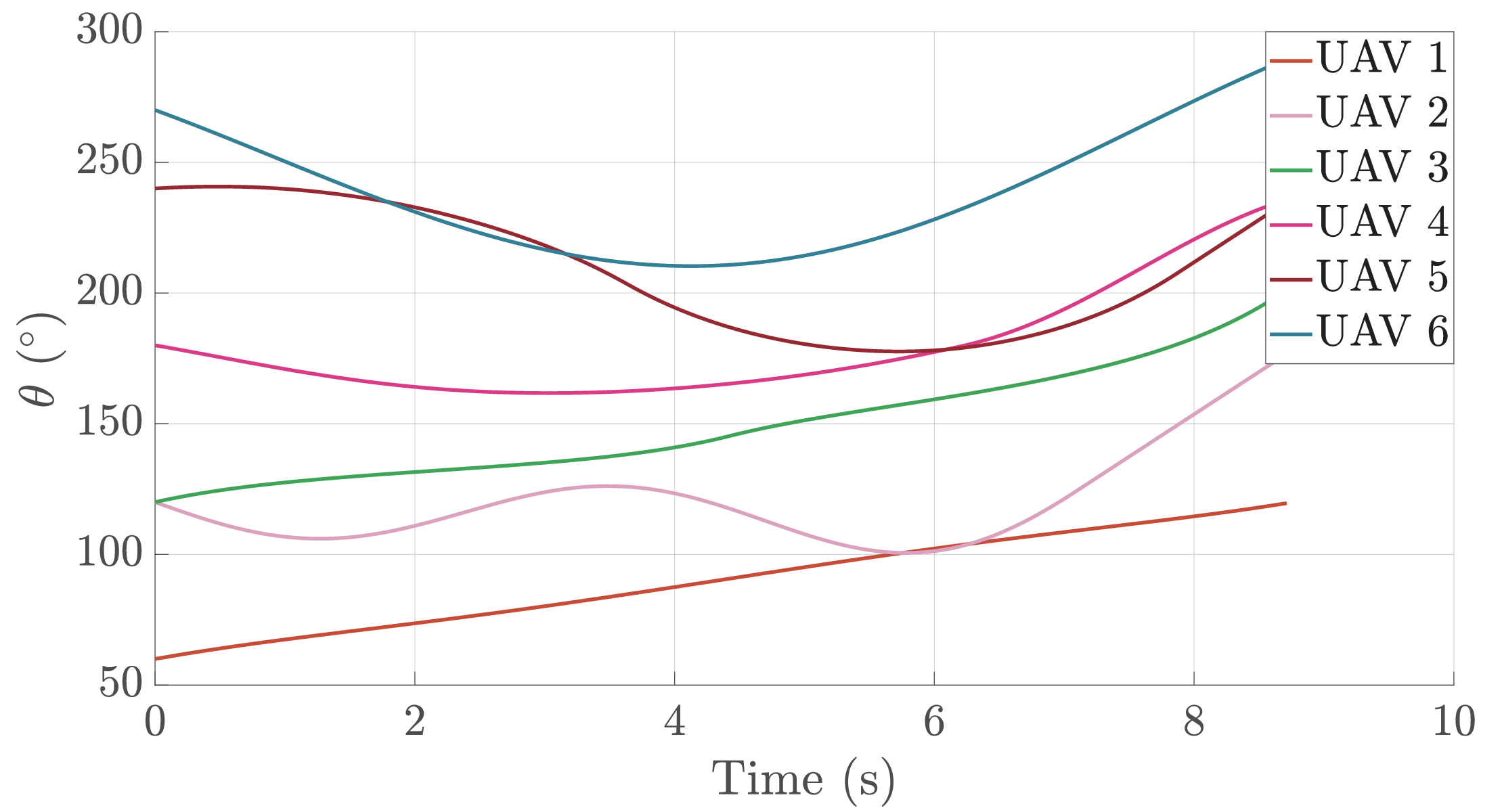}
        \caption{Time history of angular rate.}
        \label{fig8:sub2}
    \end{subfigure}
 %   \vspace{0.3cm}
    \begin{subfigure}[b]{0.31\textwidth}
        \centering
        \includegraphics[height=4cm]{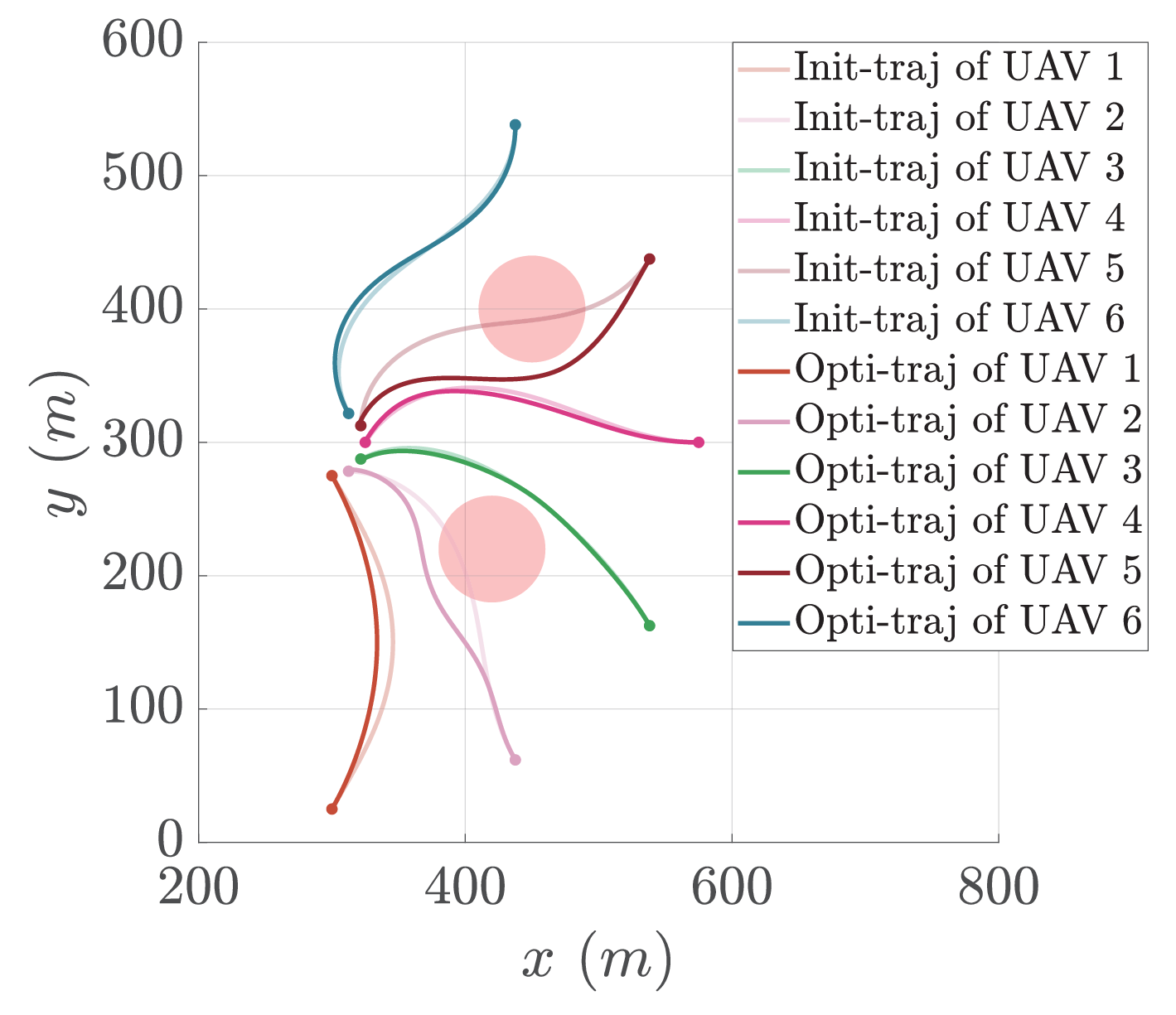}
        \caption{Trajectory.}
        \label{fig8:sub3}
    \end{subfigure}
    \caption{Distributed spatial-temporal joint optimization results
($S=2$, $\tau=20$).}
    \label{fig8}
\end{figure}

\begin{table}[h]
\caption{Optimization results for the scenario.}\label{tab3}%
\begin{tabular}{@{}lll@{}}
\toprule
Algorithms & Flight times & Optimization times\\
\midrule
CA-ADTO ($S=2$, $\tau=10$)    & $\mathrm{9.17s}$   & $\mathrm{206.41s}$  \\
CA-ADTO ($S=N/2$, $\tau=10$)    & $\mathrm{9.17s}$   & $\mathrm{231.20s}$  \\
CA-ADTO ($S=2$, $\tau=20$)    & $\mathrm{9.12s}$   & $\mathrm{308.31s}$  \\
\botrule
\end{tabular}
\end{table}

The optimization results of the UAV swarm in the scenario are shown in Figure \ref{fig7} and Figure \ref{fig8}. From Figure \ref{fig8}, we can easily see the optimization results of the CA-ADTO algorithm under a fixed bounded delay and a partial barrier set to $S=N/2$. Figure \ref{fig8} shows the corresponding results when the partial barrier remains unchanged and the bounded delay is set to $\tau=20$. Together with the optimization results of Section \ref{sec4:sub2} (Figure \ref{fig3}), the comparative data for these three different test conditions in the scenario are summarized in Table \ref{tab3} (taking the optimization results of UAV $5$ as an example). In summary, a larger partial barrier leads to longer optimization time but improved performance, while a larger bounded delay results in longer optimization time and deteriorated performance.

\section{Conclusions}\label{sec5}

In this paper, a distributed trajectory optimization method with asynchronous communication is proposed for UAV swarms under constrained communication resources. The characteristics of parameters that influence the effectiveness of the algorithm are investigated: specifically, a larger partial barrier results in a longer iteration time and better optimization performance, while a lower probability of communication connection and a larger bounded delay lead to a longer iteration time and worse optimization outcomes. The algorithm is capable of generating safe (both internally and externally) trajectories that satisfy different terminal time constraints, including sequential constraints, simultaneous arrival, and optimal idle terminal time. The validation results for various scenarios demonstrate the flexibility, convergence, and optimization capabilities of the proposed approach.

\section*{Funding Declaration}
This work was supported by the National Natural Science Foundation of China (Grant Number 52302449). The author S.H. has received research support from the National Natural Science Foundation of China.

\section*{Data Availability Statement}
The data that support the findings of this study are available from the corresponding author, S.H., upon reasonable request.

\bibliography{references}

\end{document}